\PassOptionsToPackage{table}{xcolor}
\documentclass[letterpaper,10pt]{article}
\usepackage{tabularx} 
\usepackage{amsmath}  
\usepackage{graphicx} 
\usepackage[margin=1in]{geometry} 
\usepackage{cite} 
\usepackage[final]{hyperref} 
\hypersetup{
	colorlinks=true,       
	linkcolor=blue,        
	citecolor=blue,        
	filecolor=magenta,     
	urlcolor=blue         
}
\usepackage{blindtext}
\usepackage{natbib}
\usepackage[utf8]{inputenc} 
\usepackage[T1]{fontenc}    
\usepackage{url}            
\usepackage{booktabs}       
\usepackage{amsfonts}       
\usepackage{nicefrac}       
\usepackage{microtype}      
\usepackage{mathrsfs}
\usepackage{dsfont}
\usepackage{amssymb}  
\usepackage{makecell}
\usepackage{amsmath}
\usepackage{multirow}
\usepackage{booktabs}
\usepackage{algorithm}
\usepackage{algpseudocode}
\usepackage{amsthm}
\theoremstyle{definition}

\usepackage{xspace}
\usepackage{longtable}
\usepackage{wrapfig}
\usepackage{enumitem}
\usepackage{etoc}
\setlist{leftmargin=5mm}
\usepackage[export]{adjustbox}

\usepackage{todonotes}
\usepackage{times}

\usepackage{titlesec}
\usepackage{makecell}
\usepackage[table]{xcolor}
\usepackage{wasysym}  
\usepackage{pifont}   
\usepackage{subcaption}
\usepackage{graphicx}
\usepackage{array}
\usepackage{multirow}
\usepackage{multicol}
\usepackage{etoc}
\usepackage{listings}

\newcommand{\dataset}{CultureVerse\xspace}
\newcommand{\method}{CultureVLM\xspace}
\newcommand{\vlms}{VLMs\xspace}

\newcommand{\llava}{\textsc{LLaVA-v1.5}\xspace}
\newcommand{\llavanext}{\textsc{LLaVA-NeXT-v1.6}\xspace}
\newcommand{\llavaone}{\textsc{LLaVA-OneVision}\xspace}
\newcommand{\llama}{\textsc{LLaMA}\xspace}

\definecolor{cvprblue}{rgb}{0.21,0.49,0.74}
\definecolor{my_green}{RGB}{51,102,0}
\usepackage{tcolorbox}
\definecolor{myboxcolor}{RGB}{245,245,245} 
\definecolor{myframe}{RGB}{0,0,128}
\newtcolorbox{mybody}{
  colback=myboxcolor,
  colframe=myframe,
  boxrule=1pt, 
  left=1pt,
  right=1pt,
  top=1pt,
  bottom=1pt,
}
\newcommand{\green}[1]{\textcolor{my_green}{#1}}

\title{\method: Characterizing and Improving Cultural Understanding of Vision-Language Models for over 100 Countries}
\author{%
Shudong Liu$^{1}$\footnote{Contact: nlp2ct.shudong@gmail.com.}, Yiqiao Jin$^{2}$, Cheng Li$^{3}$, Derek F. Wong$^1$\\ Qingsong Wen$^4$, 
Lichao Sun$^5$, Haipeng Chen$^6$, Xing Xie$^3$, Jindong Wang$^6$\footnote{Corresponding author: jwang80@wm.edu.}\\
    \textsuperscript{1}University of Macau \quad 
    \textsuperscript{2}Georgia Institute of Technology \quad \textsuperscript{3}Microsoft Research
    \\ 
    \textsuperscript{4}Squirrel AI \quad 
    \textsuperscript{5}Lehigh University\quad 
    \textsuperscript{6}William \& Mary
\\ \textcolor{magenta}{\large\url{https://culturevlm.github.io}}
}

\date{}

\begin{document}
\maketitle

\begin{abstract}
Vision-language models (\vlms) have advanced human-AI interaction but struggle with cultural understanding, often misinterpreting symbols, gestures, and artifacts due to biases in predominantly Western-centric training data. 
In this paper, we construct \dataset, a large-scale multimodal benchmark covering $19,682$ cultural concepts, $188$ countries/regions, $15$ cultural concepts, and $3$ question types, with the aim of characterizing and improving \vlms' multicultural understanding capabilities. 
Then, we propose \method, a series of \vlms fine-tuned on our dataset to achieve significant performance improvement in cultural understanding. 
Our evaluation of $16$ models reveals significant disparities, with a stronger performance in Western concepts and weaker results in African and Asian contexts. Fine-tuning on our \dataset enhances cultural perception, demonstrating cross-cultural, cross-continent, and cross-dataset generalization without sacrificing performance on models' general VLM benchmarks. 
We further present insights on cultural generalization and forgetting.
We hope that this work could lay the foundation for more equitable and culturally aware multimodal AI systems.
\end{abstract}

\addtocontents{toc}{\protect\setcounter{tocdepth}{-1}}
\section{Introduction}
Vision-language models (\vlms) have achieved great performance in various tasks, such as visual question answering and captioning~\citep{GPT4V,hurst2024gpt,team2023gemini,claude3-5,wang2024qwen2,liu2024visual}.
Meanwhile, one of the most vital aspects of human experience, cultural understanding, which encompasses language, cultural values, social norms, culinary practices, and artistic expressions--remains a challenging area for these models~\citep{winata2024worldcuisines,adilazuarda2024towards}.

\noindent \textbf{Challenges.} Cultural understanding is essential for AI systems intended for global deployment, as it enables them to interact appropriately and sensitively with users of diverse cultural, ethnic, and social backgrounds. 
However, current VLMs often struggle to grasp the deeper cultural meanings embedded in symbols and artifacts. 
For instance, a VLM may identify an eagle as merely a bird, overlooking its symbolic significance as a national emblem representing the spirit and identity of the United States. 
Similarly, the lotus flower is not only a plant, but a profound symbol of purity and spiritual enlightenment in Indian culture. 
Gestures present an even more complex challenge: the ``OK'' hand gesture, which conveys a positive meaning in North American countries, is interpreted as offensive in countries such as Brazil and Turkey~\citep{medhat2015gestures}.
Misinterpretations of culturally significant symbols can lead to misunderstandings and even cause offense. 

These challenges partially stem from inherent biases and limitations in VLMs' training data:
\emph{1) Skewed Domain Coverage.} Pre-training images and texts predominantly feature \emph{generic} daily scenes or natural settings, often lacking coverage of \emph{culturally specific} artifacts, traditions, beliefs, and historical sites. Models may fail to interpret culturally significant symbols, particularly those from underrepresented regions. 
\emph{2) English-centric Data and Western Bias. } 
The texts for pre-training VLMs is primarily sourced from English content~\citep{naous2023readme++,jin2024better}, which predominantly represents high-resource cultures, resulting in a Western Bias~\citep{young2014western,deng2024deconstructing}. This limits the models' understanding of diverse cultures, especially those in the global south~\citep{chiu2024culturalbench}.

Addressing these challenges is crucial for developing culturally aware AI systems that can engage effectively and respectfully with global users. 
Although there have been efforts to build culturally aware LLMs through data collection~\citep{shi2024culturebank,li2024culturellm,chiu2024culturalbench}, improving \vlms for cultural understanding remains in its infancy.
\vlms require multimodal input----both texts and images--making the collection or generation of culturally rich training data even more challenging, especially for low-resource cultures. 
From the \emph{benchmarking} perspective, existing datasets for VLMs~\citep{romero2024cvqa,nayak2024benchmarking} usually rely on human annotators for data curation. 
These datasets are limited in scale, often lack sufficient regional and national representation, and may not capture deep cultural relevance. 
From the \emph{modeling} perspective, there is a lack of work aimed at building culturally aware \vlms, a significant obstacle to AI equity for underrepresented cultures.

\begin{figure*}[ht]
\centering 
\includegraphics[width=.8\linewidth]{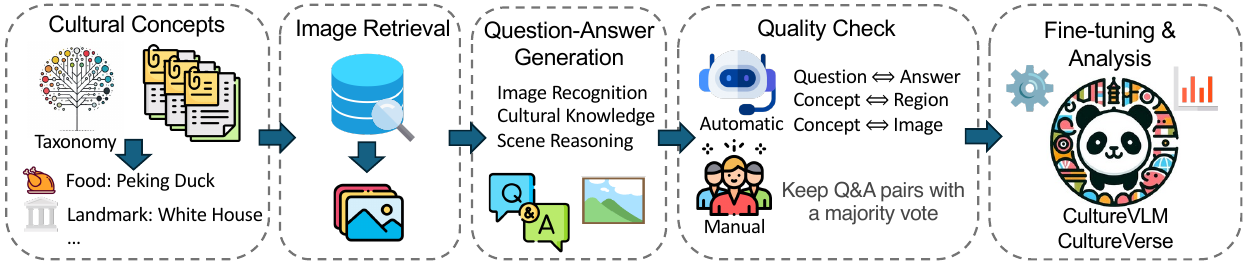}
\caption{Our pipeline to build \dataset and \method.
}
\label{data_collection}
\end{figure*}

\noindent \textbf{This Work.} We take the first step toward advancing cultural understanding in \vlms through both comprehensive benchmarking and targeted model improvements. 
To achieve this, we construct \dataset, a large-scale multimodal dataset to evaluate and enhance the multicultural capabilities of \vlms. Our flexible pipeline (\figurename~\ref{data_collection}) can easily integrate additional languages and cultures, ensuring adaptability and inclusiveness. Our work lays a solid foundation for developing more equitable AI systems that address the needs of developing nations, ethnic minorities, and underrepresented cultures. 
The key findings are as follows.

\begin{itemize}[leftmargin=1em]
\setlength\itemsep{0em}
    \item \textbf{Disparity in Cultural Understanding}: \emph{all} \vlms show highly consistent regional disparities in cultural understanding, with the highest cultural understanding for the Americas, followed by Europe and Oceania, and the weakest understanding for Asia and Africa.
    \item \textbf{Training for Enhanced Cultural Perception}: Fine-tuning effectively enhances the cultural perception of VLMs, narrowing the gaps in cultural understanding across different regions and categories without significantly compromising the model's general capabilities.
    \item \textbf{Model and Data Scale Enhance Cultural Understanding}: Cultural understanding is generally positively correlated with model size, though not absolutely, as demonstrated by the Llama 3.2-11B model achieving performance comparable to that of Qwen 2-72B. Regarding fine-tuning, larger training datasets lead to more significant improvements.
    \item \textbf{Generalization across Cultures, Concepts, Continent, and Datasets}: Due to the inherent correlations between cultures of different regions and types, fine-tuning for cultural understanding exhibits reasonable generalization across different cultures, concepts, continents, and even datasets, showing great potential to improve cultural understanding via generalization research.
\end{itemize}

\noindent \textbf{Contributions.} Our contributions are as follows:
\begin{itemize}[leftmargin=1em]
\setlength\itemsep{0em}
    \item \textbf{Large-scale Dataset.} We present \dataset, a massive scale benchmark consisting of 19,682 cultural concepts and 228,053 samples, covering 3 tasks, 188 countries or regions, and 15 cultural concepts. The test set includes 11,085 widely recognized concepts and their corresponding 31,382 samples.
    \item \textbf{Comprehensive Evaluation.} We evaluate a wide range of cultural concepts across 16 open-source and proprietary models of varying scales.
    \item \textbf{Improvement of Multimodal Cultural Understanding.} We present \method, which includes a flexible and cost-effective data collection and construction process and a series of VLMs fine-tuned on our dataset. Experimental results demonstrate that our models enhance cultural understanding while maintaining general capabilities and exhibiting a degree of generalization abilities.
\end{itemize}

\section{Related Work}

\textbf{Cultural Bias in LLMs and VLMs.} 
Recent research has increasingly focused on cultural biases present in large language models (LLMs).
\citet{johnson2022ghost} investigated conflicts between model outputs and input values and found that GPT-3’s responses often aligned more closely with dominant U.S. cultural norms.
Similarly, \citet{naous2023having} observed a bias toward Western cultural perspectives in models processing Arabic text.
The Cultural Alignment Test (CAT), based on Hofstede’s cultural dimensions framework \citep{lamport94}, was used to evaluate the cultural alignment of models like ChatGPT and Bard across various regions, showing that GPT-4 exhibited the strongest alignment with U.S. values~\citep{masoud2023cultural}.
Additionally, \citet{cao2023assessing} found that, while ChatGPT was well-aligned with American cultural values, it struggled to represent other cultures accurately, especially when responding to English prompts. \citet{liu2023multilingual} further reported that multilingual LLMs showed limited proficiency in reasoning with proverbs and revealed a ``culture gap'' in translation tasks~\citep{liu2023knn}.

\noindent \textbf{Datasets and Models for Cultural Understanding.}
Most of the research used existing datasets as cultural datasets.
\citet{wang2023not} introduced a benchmark based on the World Values Survey (WVS) \citep{wvs} and the Political Culture and Trust (PCT) dataset \citep{mudde20162012}.
Subsequent works include the Cultural Alignment Test \citep{masoud2023cultural}, NORMSAGE \citep{fung2022normsage}, WorldValueBench \citep{zhao2024worldvaluesbench}, and NORMAD \citep{rao2024normad}, each drawing on various existing datasets.
Other sources include CultureAtlas \citep{fung2024massively} and MAPS~\citep{liu2023multilingual}, which collected data from Wikimedia, while Candle~\citep{nguyen2023extracting} and CultureBank~\citep{shi2024culturebank} gathered data from social media platforms, including TikTok and Reddit. In contrast, there is a growing trend toward automatic data augmentation such as \citep{li2024culturellm, li2024culturepark}.
A strand of research focuses on training cultural-specific LLMs by assembling large-scale pre-training datasets, followed by fine-tuning to enhance alignment~\citep{pires2023sabia,chan2023harmonizing,nguyen2023seallms,pipatanakul2023typhoon,abbasi2023persianllama,lin2023taiwan}. 
Instead of relying on massive data collection, \citet{li2024culturellm,li2024culturepark} proposed cost-efficient approaches to fine-tuning cultural-specific models by data augmentation.

Unlike LLMs, training data are significantly more difficult to obtain for \vlms. The research in \vlms' cultural bias is still preliminary, with most efforts in \emph{manual} data collection~\citep{liu2021visually,romero2024cvqa,nayak2024benchmarking,bhatia2024local}. 
MaRVL~\citep{liu2021visually} introduced a protocol for building an ImageNet-style hierarchy that represents a wider range of languages and cultures. CVQA~\citep{romero2024cvqa} proposed a culturally diverse multilingual visual question-answering benchmark designed to encompass a wide variety of languages and cultural contexts, engaging native speakers and cultural experts in the data collection process.
CulturalVQA~\citep{nayak2024benchmarking} developed a visual question-answering benchmark focused on evaluating \vlms’ understanding of culturally diverse, geographically specific content.
GlobalRG~\citep{bhatia2024local} presents two challenging tasks: retrieval across cultural universals and culturally specific visual grounding.

However, these datasets are often limited in size, lack sufficient regional and national representation, and may exhibit weak cultural relevance. More critically, at the model level, there has been no significant effort to develop culturally aware VLMs, posing a major barrier to AI equity for underrepresented cultures.
\tablename~\ref{data_compare_transposed} shows the key difference between our benchmark and existing multimodal cultural benchmarks, clearly showing that our benchmark contains more diverse and large-scale data.

\begin{table}[t!]
    \centering
    \small
    \begin{tabular}{l@{\hskip 3pt}c@{\hskip 3pt}c@{\hskip 3pt}c@{\hskip 3pt}c@{\hskip 3pt}c}
        \toprule
        Dataset & Country & Concept & Image & Question & Multi. \\
        \midrule
        MaRVL \citep{liu2021visually} & 5 & 454 & 4,914 & 5,670 & \ding{55} \\
        CVQA \citep{romero2024cvqa} & 28 & - & 4,560 & 9,044 & \ding{55} \\
        CulturalVQA \citep{nayak2024benchmarking} & 11 & - & 2,328 & 2,328 & \ding{55} \\
        GlobalRG \citep{bhatia2024local} & 50/15 & - & 3,591 & - & \ding{51} \\
        \midrule
        \dataset\ & 188 & 11,085 & 11,085 & 31,382 & \ding{51} \\
        \hfill w/ Train Set & 188 & 19,682 & 74,959 & 196,673 & \ding{51}\\
        \bottomrule
    \end{tabular}
    \caption{Comparison of various cultural datasets with features. `Multi.' indicates whether the dataset provides multi-faceted questions rather than just one single type. }
    \label{data_compare_transposed}
\end{table}

\section{\dataset: A Scalable Benchmark for VLM Cultural Understanding}
\label{sec-data}
Collecting reliable large-scale culture datasets presents two key challenges: \emph{Diversity} and \emph{Scalability}. 
Achieving comprehensive coverage is especially difficult for \emph{culturally diverse} topics, particularly underrepresented groups in the Global South.
The construction and annotation of such datasets would typically require substantial human expertise from various countries and ethnic groups, resulting in poor scalability and high costs. 
Existing cultural benchmarks for \vlms usually lack adequate representation of diverse regions and communities, and often reflecting a bias towards dominant cultures~\citep{liu2021visually,romero2024cvqa,oh2024uniguard}.

To overcome these limitations, we introduce a scalable data collection pipeline that integrates automated web crawling for \emph{scalability} and \emph{diversity} with expert human annotation for \emph{reliability}.
As shown in \figurename~\ref{data_collection}, our pipeline consists of three stages: tangible cultural concept collection, question-answer generation, and quality assurance.
This hybrid approach ensures that our dataset captures a wide spectrum of cultural contexts while maintaining high standards of data quality and relevance.

\subsection{Tangible Cultural Concept Collection}
\label{sec:concept_collection}
To construct a comprehensive set of cultural concepts, a common approach is to employ a bottom-up strategy that retrieves specialized knowledge from open web documents.
For example, ~\citet{fung2024massively} begin with an initial set of cultural topics (e.g., education, marriage customs, and holiday traditions), collect relevant Wikipedia documents, and expand their scope through linked connections. 
However, many resulting documents primarily describe general, abstract, or high-level concepts, such as \texttt{Renaissance Art} or \texttt{Mediterranean cuisine},
which often lack specific, unambiguous visual representations. 

\noindent \textbf{Concept Construction.} To overcome this issue, we adopt a top-down approach, starting with $15$ predefined categories of tangible cultural concepts such as food, festivals, landmarks, and performing arts, as shown in \tablename ~\ref{tb-concept-category}. 
These categories were chosen to capture culturally distinctive and visually recognizable elements suitable for image retrieval and analysis. 
We then use GPT-4o to process all relevant Wikipedia documents, extracting conceptual entities that align with the $15$ predefined categories. 
To ensure the quality and specificity of the extracted entities, we implement a $3$-step filtering process on the extracted conceptual entities: 
1) \emph{Entity Consolidation.} We unified duplicate entities, merging those that are identical or differ solely by case, and eliminated entities with formatting issues or irregularities;
2) \emph{Frequency-based Thresholding.} We retained only entities that appeared at least twice across the documents from a given country, ensuring that the concepts are well-recognized within their cultural context;
3) \emph{Entity Refinement.} We filtered out overly abstract or generic entities such as \texttt{Imperial Cuisine} and those lacking distinct regional specificity such as \texttt{Steak}
using additional judgment by GPT-4o. 
Through this refined process, we curated a collection of over $19,682$ cultural concepts from $188$ countries, as shown in Table~\ref{tb-country-concept}. 
Our pipeline ensures that the selected concepts are diverse and well-recognized, relevant for evaluating the capabilities of \vlms in understanding global cultural diversity. 

\noindent \textbf{Image Retrieval.} Using these concepts and their corresponding countries, we scrape images from Google Images for each cultural concept\footnote{All images are only used for research purpose.}, obtaining five images for each concept. The first image was reserved for the test set and for human quality assessment, while the remaining four images were used for the training set. Images larger than $10$MB were compressed to ensure compatibility with typical input requirements of VLMs.

\subsection{Question-Answer Generation}
\label{Question-Answer Generation}
We designed three levels of VQA tasks to assess and improve the multicultural knowledge of \vlms:

\noindent \textbf{Image Recognition Questions} 
evaluate models' ability to identify cultural concepts in images. Accurate identification of such concepts is fundamental to retrieving relevant cultural knowledge. 
Given an image and a cultural concept, models answer questions like ``What dish is in the image?''

\noindent \textbf{Cultural Knowledge Questions} further evaluate model’s deeper understanding of the cultural background associated with the concepts. 
For each concept, we generated comprehensive descriptions, including aspects like location, characteristics, history, and cultural significance.
Subsequently, we instruct GPT-4o to formulate a question based on the introduction and the image to probe this cultural knowledge without directly naming the concept in the question. 
These questions require the model to identify cultural concepts and apply various levels of reasoning, drawing on relevant cultural knowledge to provide accurate answers. 

\noindent \textbf{Scene Reasoning Questions} are designed to assess the model’s ability to interpret, interact, and respond within culturally specific contexts, rather than simply recalling factual information as in the previous two categories.  
We curate scenarios with cultural elements or characteristics depicted in the images, providing context cues that challenge the \vlms to make contextually appropriate choices.
Using the detailed introductions of the previous steps, we prompt GPT-4o to generate scenario-based reasoning questions. 
These questions require the model to not only recognize cultural concepts but also apply contextual reasoning based on the associated cultural knowledge.

\subsection{Quality Assurance}
To ensure the integrity of the dataset, we conduct a comprehensive manual quality check on each cultural concept, along with its corresponding images and questions. Specifically, we employ human annotators (details are shown in Appendix~\ref{sec-app-human}) to inspect three main components:
\begin{itemize}
    \item \textbf{Image-Concept Alignment.} We assessed whether each cultural concept accurately represents the culture of its respective country or region and is either unique to or widely recognized within that. We started with frequency analysis and leverage GPT-4o for preliminary screening, effectively filtering out less desirable data and significantly reducing the manual review workload.
    \item \textbf{Image Quality Check.} We checked the image quality and ensured that the cultural concept is accurately presented in the image. 
    \item \noindent \textbf{Question \& Answer Validation.} 
    We verified that all three generated questions are reasonable, clear, logically sound, and have a single correct answer. Annotators refined the questions and answer options by removing redundant information and resolving any ambiguities to maintain clarity and accuracy. 
\end{itemize}

Following the quality assurance process, we utilized human annotations for the evaluation set of \dataset while applying the automated annotation pipeline to the larger training set.  With over $98$\% of the evaluation set samples correctly annotated by the automated process, we conclude that the pipeline is highly effective. Any remaining erroneous or challenging samples that could not be refined were filtered out to maintain the dataset's high quality.
Additional details on annotator accuracy are in Appendix~\ref{sec-app-acc-human}.

\begin{figure*}[t]
    \centering
    \includegraphics[width=.9\linewidth]{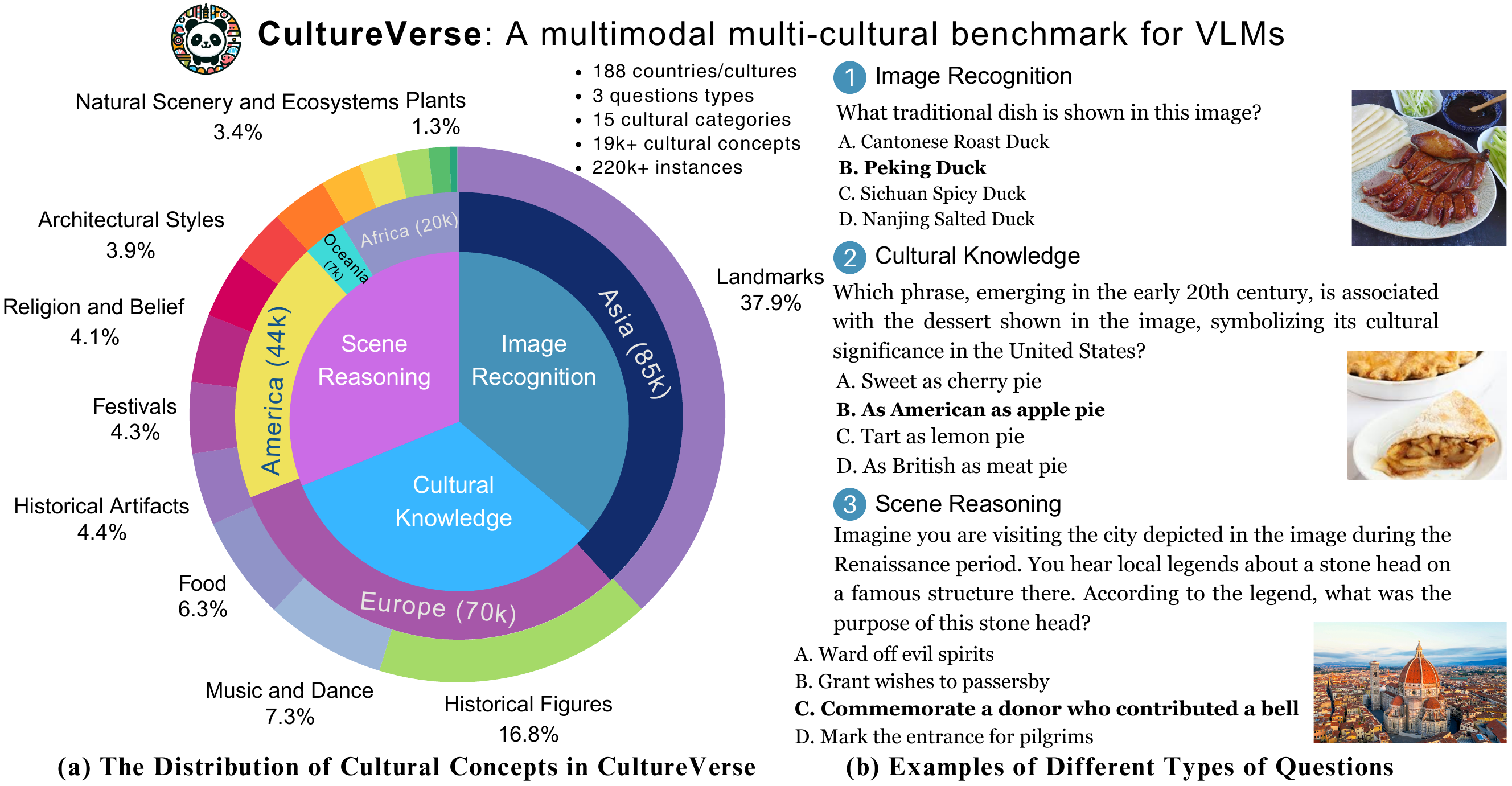}
    \vspace{-.1in}
    \caption{Overview of \dataset. In total, there are over 220k instances and 19k cultural concepts for training and evaluation, respectively, composed of 3 different types of questions from 188 countries.}
    \label{fig-dataset}
    \vspace{-.1in}
\end{figure*}

\subsection{Scalability}
Our approach to constructing multimodal cultural datasets is notably more scalable and comprehensive than existing methods. Existing ones mostly rely on manual efforts to search for cultural concepts, retrieve images, and formulate questions, significantly increasing human efforts beyond quality check \citep{chiu2024culturalbench,romero2024cvqa,jin2024mm}. 
This manual process typically results in limited or biased coverage due to the limited scope of cultural concepts explored. 
For example, some datasets \citep{bhatia2024local} cover only a dozen countries, with approximately 40 cultural concepts for each country.

In contrast, our work is the first to advance the number of specific cultural concepts to the scale of tens of thousands and to extend coverage to approximately two hundred countries/regions, as shown in Appendix~\ref{app:dataset_details}.
Additionally, our dataset construction process allows for further large-scale expansion, including the retrieval of more images and the synthesis of various types of QAs, such as open-ended QAs, multiple-choice QAs, and reasoning QAs. Our work provides a potential data source for enhancing the multicultural knowledge of VLM models.

\section{Analysis of \dataset}

\figurename~\ref{fig-dataset}(a) illustrates the distribution of three tasks (Section \ref{Question-Answer Generation}), 5 continents (188 countries, with North and South America combined into America) and 15 cultural topics in \dataset.
Since cultural concepts are collected at the country level, regions with more countries, such as Asia and Europe, naturally yield larger datasets. Detailed counts of countries and concepts are provided in the Appendix \ref{tb-country-concept}. Detailed statistics of the concepts are provided in Table \ref{data_compare_transposed}. Compared to recent multi-model culture benchmarks, \dataset is driven by tangible, presentable cultural concepts, achieving an order-of-magnitude increase in the number of countries, images, and questions. This expansion advances multimodal and multi-cultural research beyond a limited set of countries, moving toward truly global, inclusive cultural interaction and integration.
\figurename~\ref{fig-dataset}(b) shows examples of the three questions associated with one concept, and it is obvious that different types of questions aim to evaluate different abilities.

\begin{figure*}[t!]
    \centering
    \includegraphics[width=0.9\linewidth]{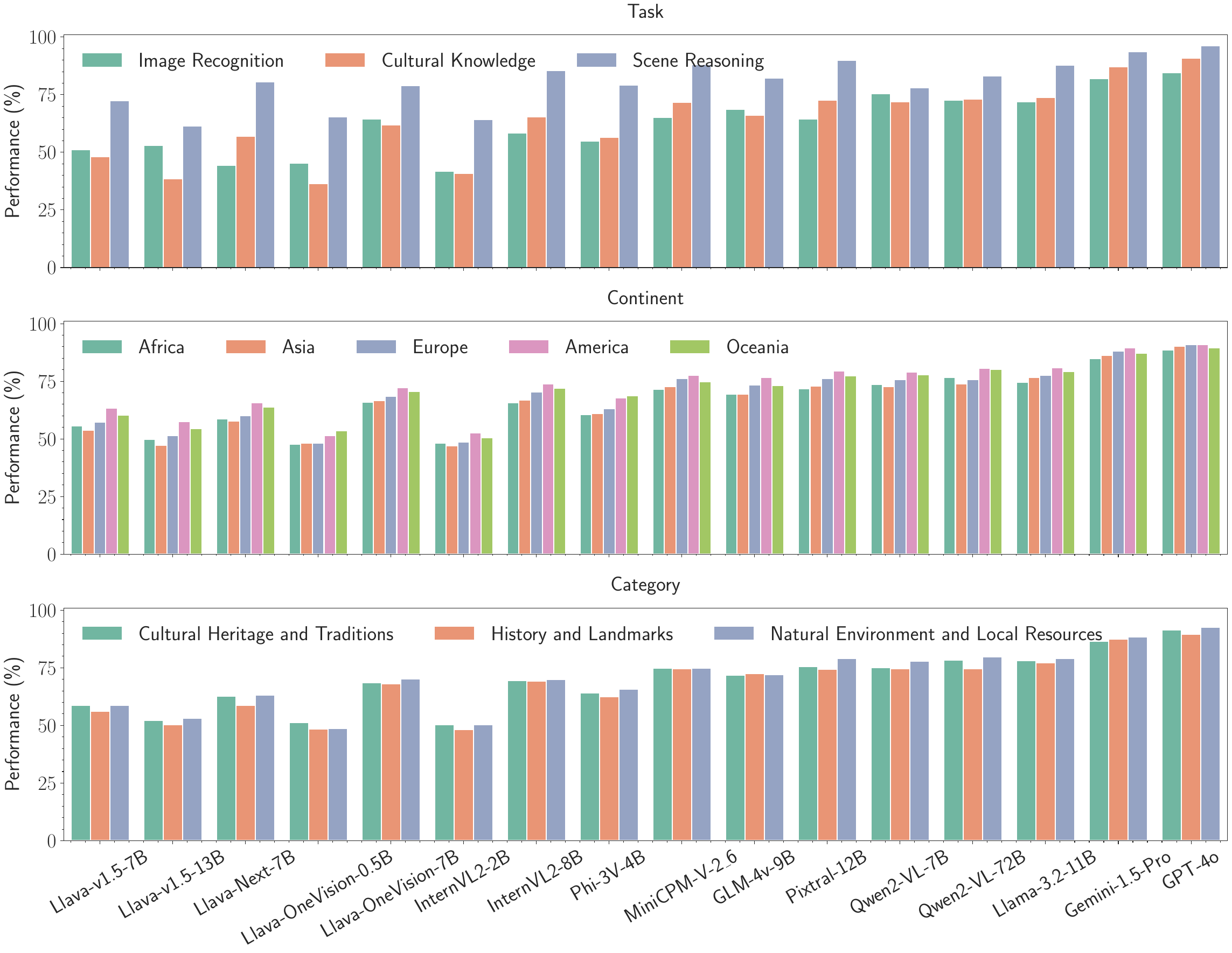}
    \vspace{-.1in}
    \caption{Accuracy of different models on three tasks (upper), five regions (middle), and three categories of concepts (lower).}
    \label{fig:main_model_performance}
    \vspace{-.1in}
\end{figure*}

\section{Experiments with \dataset}

\subsection{Experimental Setup}

\textbf{Data Split.}
To ensure a robust evaluation, we partition \dataset into training and test sets for all countries/regions, allowing us to assess transferability between regions. We select more common cultural concepts from the entire dataset for the test set, which underwent manual quality checks, while the training set includes all cultural concepts. We ensured that the images in the training and test sets did not overlap to prevent data leakage. More details can be found in Appendix \ref{app:experiment_setup}.

\noindent \textbf{Models and Hyperparameters.}
We evaluate our benchmark on $14$ open-source and $2$ proprietary VLMs. 
For multiple-choice questions, we employ greedy search decoding for deterministic predictions. We use vllm~\citep{kwon2023efficient} and lmdeploy~\citep{2023lmdeploy} toolkits to speed up inference. 
We report accuracy, in line with previous works~\citep{liu2024visual}. More details of the experiment can be seen in Appendix~\ref{app:experiment_setup}.

\subsection{Main Results}
\label{sec-app-main-result}

The main results are in \figurename~\ref{fig:main_model_performance} and detailed results are in Appendix~\ref{app-results-main}.
Our main findings are as follows.

\noindent \textbf{Task Difficulty: Cultural Scene Reasoning Outperforms Recognition.} 
From the task perspective, we observe that image recognition and cultural knowledge questions pose challenges comparable to \vlms. \emph{Image recognition} tests \vlms' ability to identify culturally specific objects or concepts, which relies heavily on diverse and relevant image data. 
For instance, recognizing traditional foods like \emph{kimchi} from Korea, or regional attire such as a \emph{sari} from India, requires the model to have encountered similar image-text pairs in its training data. 
In contrast, cultural knowledge questions assess the model’s understanding of broader cultural elements based on text-based training. For example, asking about the significance of a festival like \emph{Diwali} or the symbolism of a \emph{red envelope} during Lunar New Year taps into the model’s text-based memory, which tends to be richer due to the abundance of internet text data. 
Interestingly, the \emph{scene reasoning} task, which integrates images with contextual background (e.g., a Japanese tea ceremony scene or a Brazilian Carnival parade), tends to yield higher accuracy. This indicates that providing contextual information in texts and images allows the \vlms to better connect visual cues with their underlying cultural meanings, leading to improved performance. 

\noindent \textbf{Regional Disparities: Better Performance in Western Cultures.} 
A dominant \emph{regional disparity} is observed among all models: VLMs demonstrate the strongest cultural understanding of the Americas, followed by Europe and Oceania. 
This trend reflects the dominance of English data centered around the Global North, leading to a disproportionate focus on Western cultural content. 
North America's relatively homogenous cultural landscape, combined with fewer countries, contributes to better model performance. 
In contrast, Asia and Africa show significantly weaker results, likely due to the scarcity of digitized, English-language data and the high cultural diversity in these regions. 
For instance, Asia consists of many countries with distinct and complex cultural contexts, such as those from East Asia, South Asia, and Southeast Asia, both within and across nations. Although Asia has the most data in \dataset (see \figurename~\ref{fig-dataset}), the models struggle to capture the intra-regional and inter-regional cultural nuances, resulting in suboptimal performance.

\begin{figure}[t!]
    \centering
    \begin{subfigure}[b]{0.24\textwidth}
        \centering
        \includegraphics[width=\textwidth]{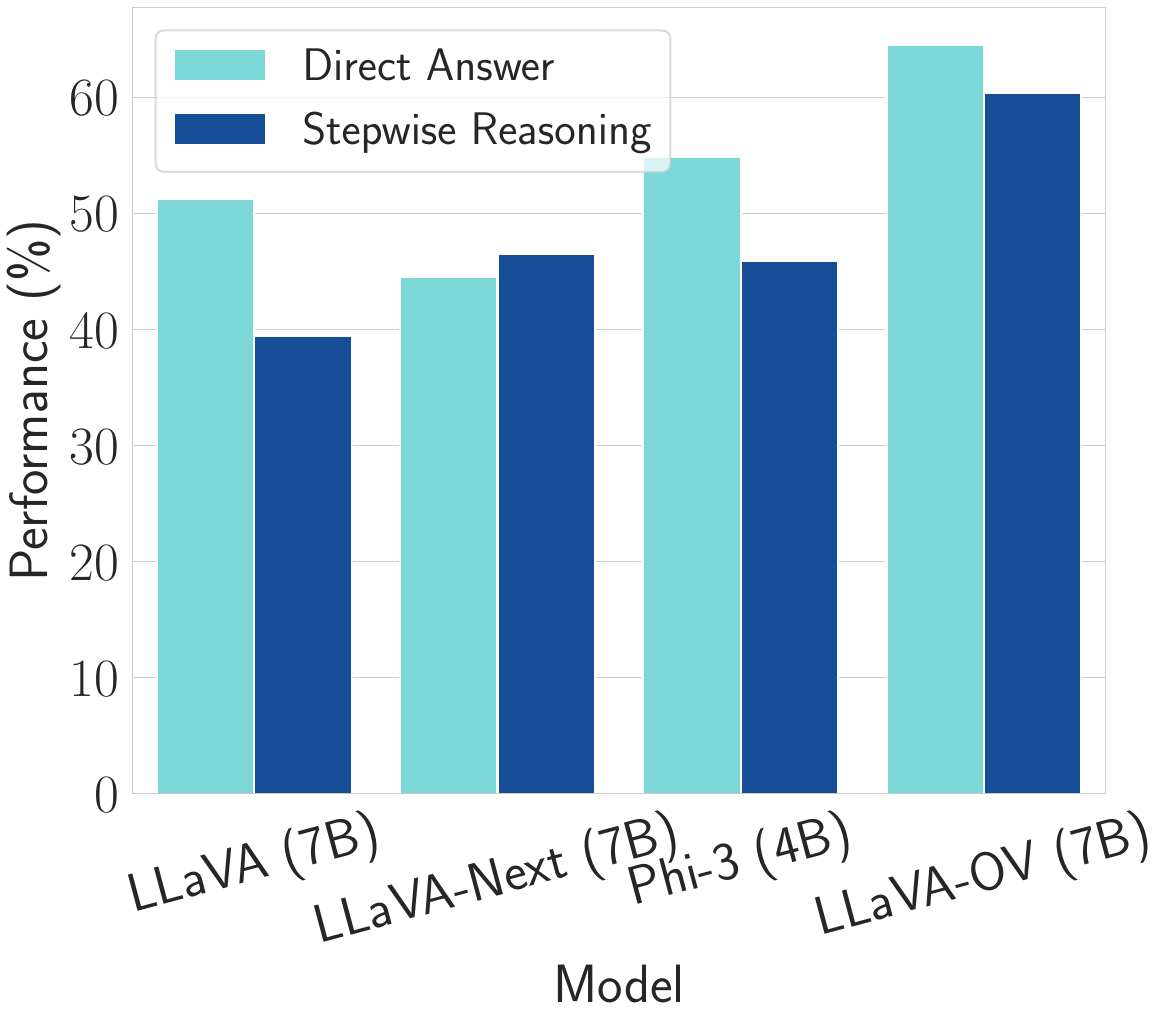}
        \caption{Direct v.s. reasoning}
    \label{fig:image_recognition_reasoning}
    \end{subfigure}
    \begin{subfigure}[b]{0.24\textwidth}
        \centering
        \includegraphics[width=\textwidth]{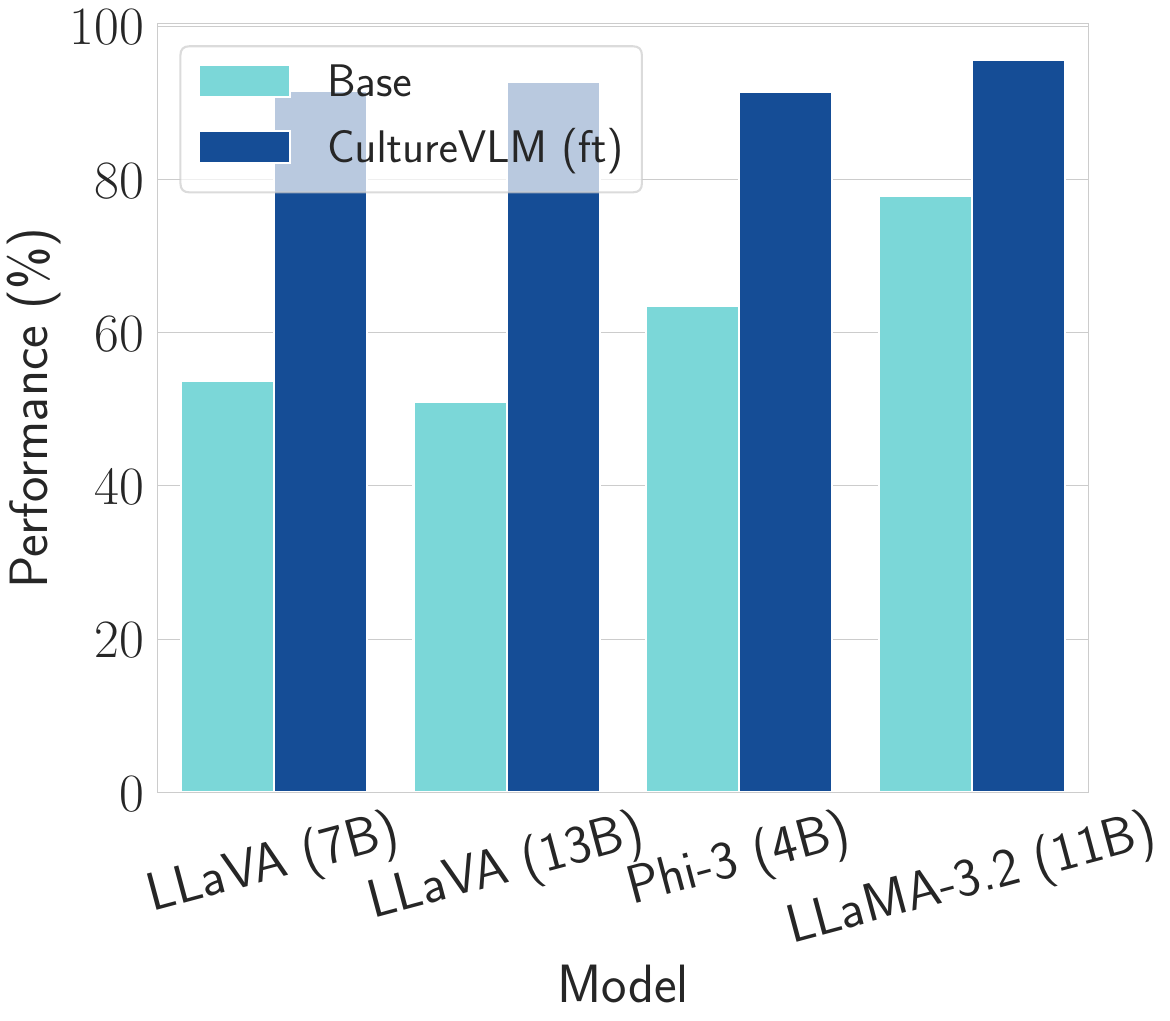}
        \caption{Fine-tuning results}
        \label{fig-fine-tune-all}
    \end{subfigure}
    \begin{subfigure}[b]{0.24\textwidth}
        \centering
        \includegraphics[width=\textwidth]{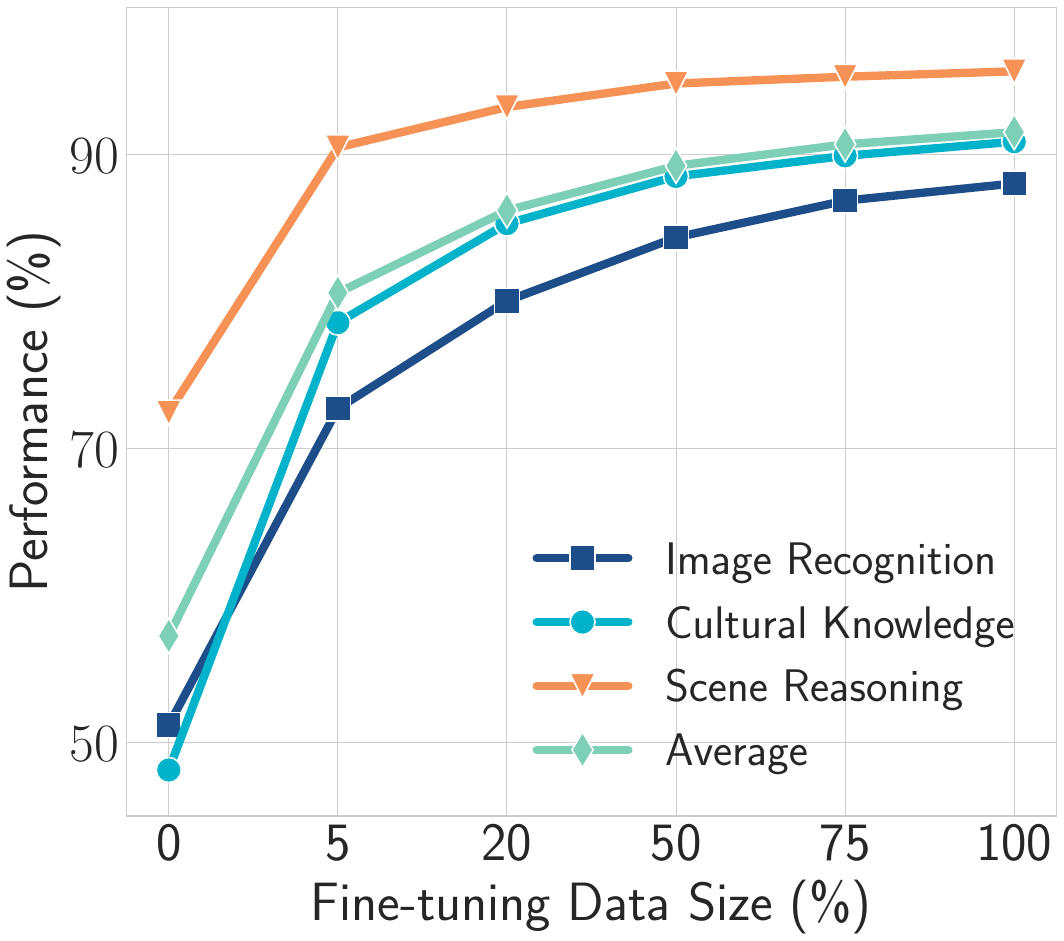}
        \caption{Tasks}
        \label{fig:fine-tune-datasize-task}
    \end{subfigure}
    \begin{subfigure}[b]{0.24\textwidth}
        \centering
        \includegraphics[width=\textwidth]{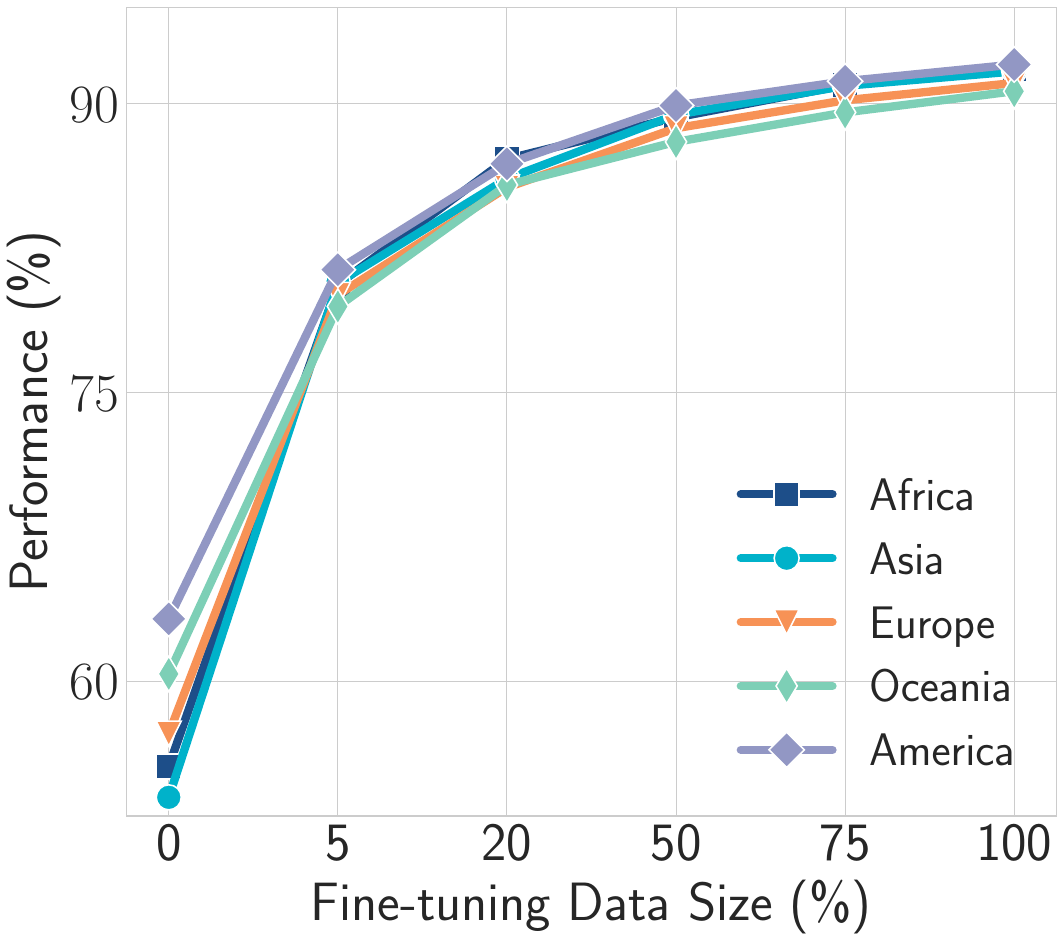}
        \caption{Continents}
        \label{fig:fine-tune-datasize-region}
    \end{subfigure}
    \caption{Results and analysis of \method by fine-tuning on our \dataset.}
    \vspace{-.1in}
    \label{fig:allfigures}
\end{figure}

\noindent \textbf{Weak Understanding of History and Landmarks.}
\vlms generally exhibit weaker recognition and understanding of cultural concepts related to history and landmarks. The primary reason is the relatively limited internet data available on historical figures and landmarks. Additionally, recognizing a landmark typically requires training data that includes images from multiple perspectives to form a comprehensive, three-dimensional understanding.

\noindent \textbf{Performance Variability from Model Level.}
Proprietary models continue to outperform open-source counterparts, with GPT-4o achieving the best results. 
Although larger models tend to demonstrate better performance, size alone is not the determining factor. 
For example, the size variations of LLaVA-1.5 and Qwen2-VL show similar performance. 
Cultural knowledge often resides in model's memory, an aspect overlooked in \vlms development. 
Thus, smaller models (e.g. Phi-3-Vision) with comparable training data can already exhibit strong cultural understanding when using similar training data as larger models. 

\begin{figure*}[t!]
    \centering
    \begin{minipage}[b]{0.25\textwidth}
        \centering
        \includegraphics[width=\textwidth]{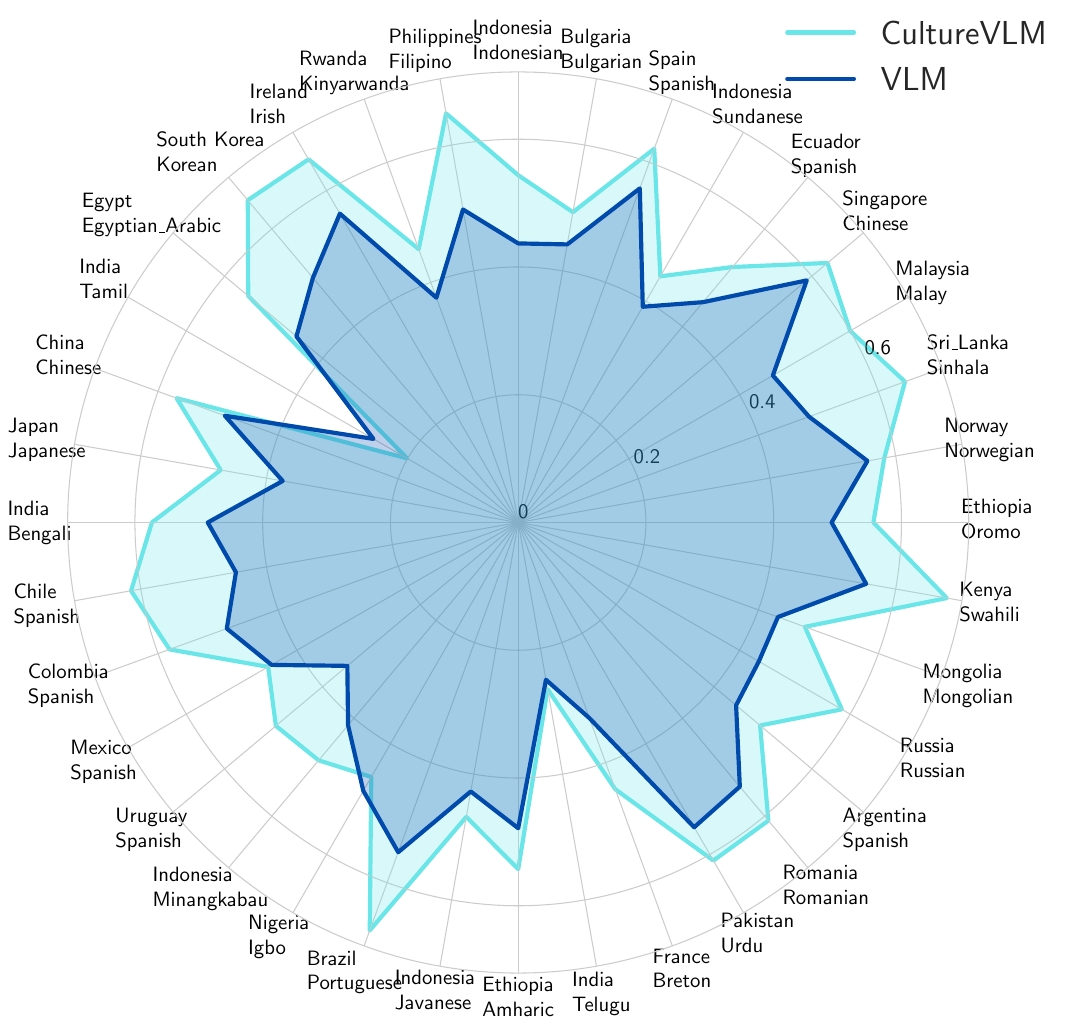}
        \caption{Results on CVQA~\citep{romero2024cvqa}.}
        \label{performance_cvqa}
    \end{minipage}%
    \hspace{0.001\textwidth} 
    \begin{minipage}[b]{0.74\textwidth}
        \centering
        \includegraphics[width=\textwidth]{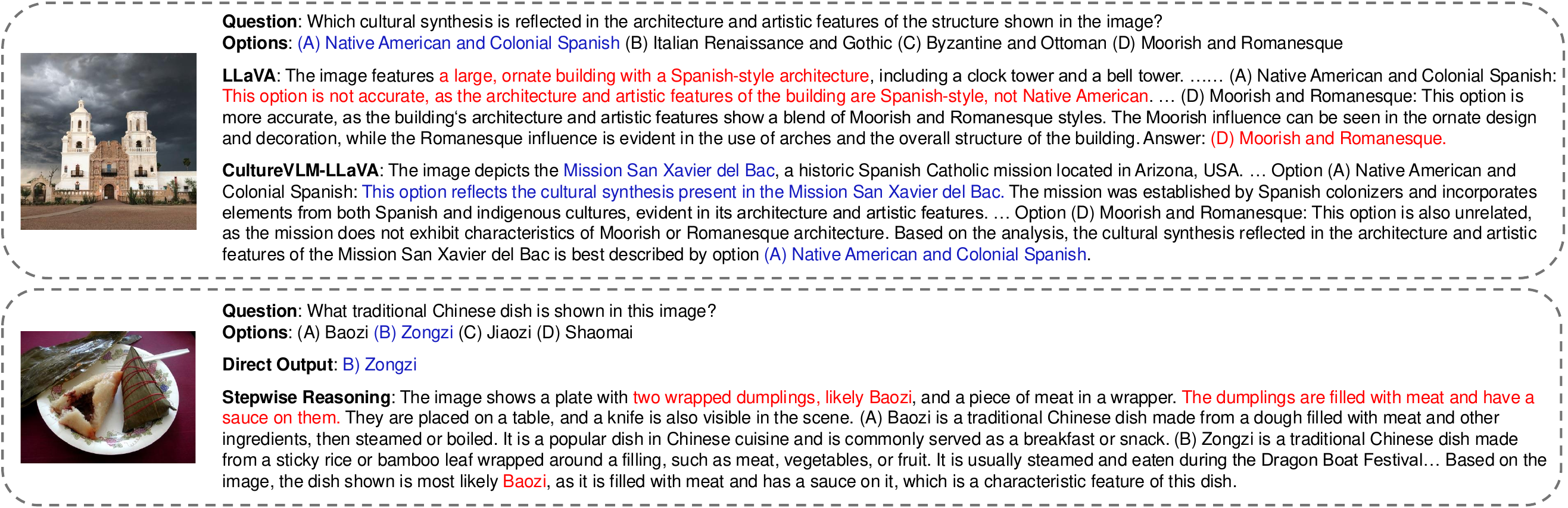}
        \caption{Case study on the effect of fine-tuning and prompt variations.}
        \label{fig:case_study}
    \end{minipage}
    \vspace{-.2in}
\end{figure*}

\noindent \textbf{Direct Answer v.s. Stepwise Reasoning.}
For image recognition tasks, we compare two prompt methods: 1) the model directly identifies and outputs the cultural concept, and 2) first provides a detailed description of the image and then analyzes and compares the options to reach a final answer. As shown in \figurename~\ref{fig:image_recognition_reasoning}, we find that stepwise reasoning does not improve cultural recognition and, in most cases, significantly \emph{impair} performance. Upon analyzing the answers, we observe that \vlms frequently exhibit hallucinations~\citep{bai2024hallucination,tonmoy2024comprehensive} during step-by-step explanations, as shown in the example in \figurename~\ref{fig:case_study}. This poor robustness suggests that while \vlms may have encountered similar images during training and associated them with certain concepts, they may lack a deeper understanding of the details and components that make up those concepts.

\subsection{Training \method}
\label{sec-exp-ablation}
We fine-tuned three models: LLaVA-1.5, Phi-3-Vision, and LLaMa-3.2-Vision, with the results presented in \figurename~\ref{fig-fine-tune-all}. Cultural knowledge, in contrast to reasoning tasks such as mathematics and coding, is relatively easier to enhance as a form of memory-based perception. Consequently, all four models achieved consistent and substantial improvements, reaching performance levels comparable to those of closed-source models.
We also performed ablation studies to analyze the impact of fine-tuning data size and option shuffling.

\begin{figure}[t!]
    \centering
\includegraphics[width=.8\linewidth]{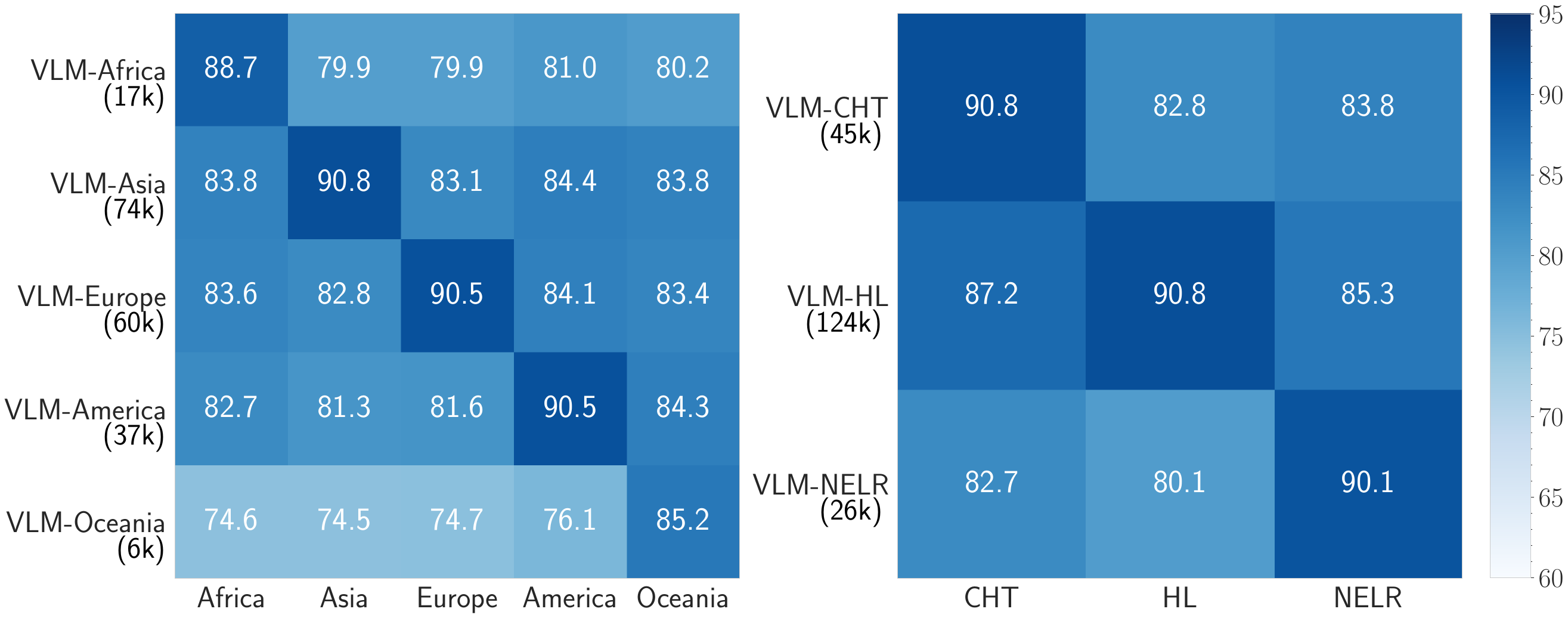}
    \caption{Generalization and Robustness. \textbf{Left}: Performance of \method (y-axis) evaluated across data from different continents (x-axis). \method achieves the highest performance for in-distribution settings, while still demonstrating strong generalizability for out-of-domain settings. \textbf{Right}: \method fine-tuned with data under different categories of \dataset (x-axis) and evaluated across various categories (y-axis). CHT denotes \emph{Cultural Heritage and Traditions}; HL denotes \emph{History and Landmarks}; and NELR denotes \emph{Natural Environment and Local Resources}.}
    \label{fig-generalization}
\end{figure}

\noindent \textbf{Impact of Data Sizes.}
We vary the number of fine-tuning examples within $[5\%, 20\%, 50\%, 75\%,100\%]$ to examine the effects of training data sizes on the final results. 
As shown in \figurename~\ref{fig:fine-tune-datasize-task} and \ref{fig:fine-tune-datasize-region}, the model's performance decreases as the training data is reduced. However, the decline is minimal, indicating that even a small amount of training data can effectively enhance the model's multicultural awareness.

\noindent \textbf{Impact of Decoding Temperatures.} We evaluate the performance of the original models and CultureVLMs under different temperature and decoding settings, as shown in Appendix \ref{app-results-finetuning}. It can be observed that VLMs perform better when the temperature is lower and decoding diversity is reduced. However, when the temperature reaches 1.0, there is a noticeable and expected decline in performance.

\subsection{Generalization and Robustness}
\label{sec:exp_regions}

To evaluate the generalizability of \vlms in multicultural contexts, we partitioned the training data by continents (Americas, Asia, Europe, Africa, Oceania) and fine-tuned LLaVA-1.5-7B~\citep{liu2024visual} separately on each subset. 
Table~\ref{tb-acc-continents} in the appendix illustrates the performance of each model trained on specific continental data and tested in all regions, and \figurename~\ref{fig-generalization} shows the aggregated results.

\noindent \textbf{Intra- and Inter-continent Generalization.}
The diagonal values represent cases where the model was trained and evaluated on data from the same continent, consistently yielding the highest scores. 
Notably, Asia achieved the best performance ($90.8$), followed closely by Europe and the Americas ($90.5$), indicating strong regional specialization. 
Furthermore, the off-diagonal values reveal models' ability to generalize across regions. 
Although cross-region scores are generally lower, the model still exhibits reasonable transferability. 
Fine-tuning on Oceania, however, resulted in the lowest average performance drop, from $85.2$ to $75.0$, suggesting a distinct data distribution for this region. 

\noindent \textbf{Robustness across Concepts.} 
We grouped the $15$ concepts into $3$ main classes: \emph{Cultural Heritage and Traditions} (CHT), \emph{History and Landmarks} (HL), and \emph{Natural Environment and Local Resources} (NELR) using GPT-4o. 
We then conducted training in each group and evaluated on all groups. The specific category mappings are in Appendix~\ref{app:dataset_details}. 
As shown in \figurename~\ref{fig-generalization} (right), models generally perform the best when trained and evaluated within the same category (in-distribution). The model fine-tuned on HL exhibits greater generalization, achieving an average performance of $87.8$, compared to NELR ($84.3$). 
High off-diagonal scores, such as CultureVLM-HL's $87.2$ when tested on CHT, indicate substantial cross-category knowledge transfer, particularly between culturally related classes. 

\noindent \textbf{Cross-dataset Generalization.}
We also evaluated the generalization ability of \method for cultural reasoning on other datasets. As shown in \figurename~\ref{performance_cvqa}, we tested LLaVA-1.5-7B on CVQA~\citep{romero2024cvqa} before and after fine-tuning. It is evident that our \method achieves improvements across most countries and cultures ($7\%$ improvement on average), which can be attributed to the comprehensive coverage of our dataset global nations and cultures, indicating potential of \dataset for cultural research.

\subsection{Catastrophic Forgetting}
Catastrophic forgetting~\citep{kirkpatrick2017overcoming} refers to the phenomenon where a model loses previously learned knowledge when trained on new information, especially when the new data diverges significantly from the pretraining data. This can be especially problematic in cultural knowledge acquisition, as it may cause the model to compromise essential commonsense knowledge in favor of culture-specific details. 

\begin{figure}
    \centering
    \includegraphics[width=0.7\linewidth]{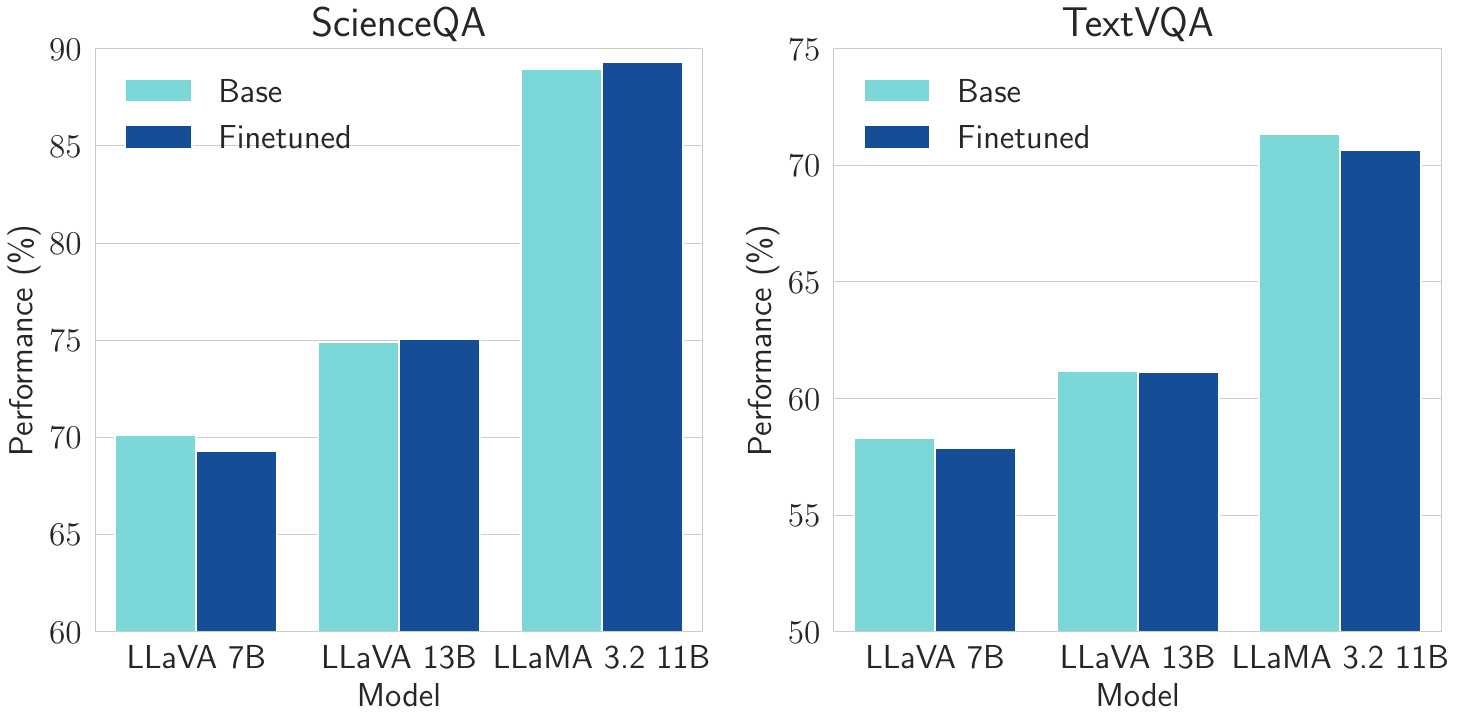}
    \caption{Performance of open-source and proprietary models on general VQA datasets. \emph{Base} refers to the original model, while \emph{Finetuned} refers to the model adapted using our \dataset. The comparable performance across both versions suggests that finetuning on our dataset preserves the models’ natural language understanding and commonsense reasoning abilities.}
    \label{fig:general_vqa}
    \vspace{-.1in}
\end{figure}

To assess this, we evaluate the models on standard VQA benchmarks, including ScienceQA~\citep{saikh2022scienceqa} and TextVQA~\citep{singh2019towards} to determine if the process of acquiring cultural knowledge inadvertently diminishes their grasp of general commonsense concepts. The results in \figurename~\ref{fig:general_vqa} reveal that our fine-tuned \method merely influences general VLM tasks, indicating the versatility of the solution.

\subsection{Case Study}
\label{exp-case}

\figurename~\ref{fig:case_study} shows the responses of LLaVA and \method. We incorporate extensive explanations into the training data, enriching \method with substantial knowledge that enhances its cultural recognition and understanding ability. More analysis and case studies are in Appendix \ref{sec-app-case-stydy}.

\section{Conclusion and Limitation}
\label{sec-conclusion}

We constructed \dataset, a comprehensive multimodal benchmark to characterize the multicultural understanding of \vlms. 
Extensive evaluation shows significant performance disparities across regions and tasks, highlighting \vlms' strong biases towards Western cultural contexts and their weaker performance in underrepresented regions like Africa and Asia. 
Using supervised fine-tuning in \dataset, we demonstrated effective enhancements in cultural perception and cross-cultural generalization. 
Our findings underscore the importance of culturally diverse training data and provide actionable insights to improve VLMs. 

We also acknowledge the following limitations.
First, while we use languages as proxies for cultural boundaries in the development of \method, we acknowledge that language alone does not capture the full complexity of culture. 
This simplification was made to address challenges in defining cultural contexts, inspired by previous research~\citep{li2024culturellm,appadurai1996modernity,myung2024blend}. 
Our pipeline and approach remain flexible to incorporate additional languages and cultures. 
Second, due to resource constraint cost, we limit our fine-tuning experiments to the current set of models. 
Evaluating a wider range of models could yield further insight. 
Third, the current \dataset only contains multiple-choice questions. Exploring open-ended questions could offer additional avenues for assessment. 
Our data and models will be released to the public after ethical reviews.

{
    \small
    \bibliographystyle{ieeenat_fullname}
    \bibliography{culturevlm}
}

\newpage

\appendix
\onecolumn
\begin{center}
    {\large \textbf{Appendix\\CultureVLM: Characterizing and Improving Cultural Understanding of\\
Vision-Language Models for over 100 Countries}}
\end{center}

\etocdepthtag.toc{mtappendix}
\etocsettagdepth{mtchapter}{none}
\etocsettagdepth{mtappendix}{subsection}

\tableofcontents

\section{Details of the \dataset Dataset}
\label{app:dataset_details}

\tablename~\ref{tb-concept-category} shows the definition of cultural concepts in our dataset.
\tablename~\ref{tb-country-concept} shows the number of concepts in different countries from the evaluation set.
In total, we have $11,085$ evaluation samples and $19,682$ training samples from $188$ countries/regions.

\begin{table*}[htbp]
    \centering
    \resizebox{.85\textwidth}{!}{
    \begin{tabular}{lll}  
        \toprule  
        \textbf{Overall Category} & \textbf{Category} & \textbf{Example Description} \\  
        \midrule  
        
        \multirow{6}{*}{\makecell{Cultural Heritage and \\Traditions}} 
        & Festivals & Unique festival celebration scenes \\  
        & Traditional Clothing & Ethnic clothing, festival attire \\  
        & Handicrafts and Artifacts & Ethnic handicrafts, traditional handmade items \\  
        & Music and Dance & Traditional musical instruments, dance scenes \\
        & Religion and Belief & Temples and churches, religious ceremonies \\
        & Entertainment and Performing Arts & Theater performances, street performers \\
        
        \midrule
        \multirow{4}{*}{History and Landmarks} 
        & Famous Landmarks & Famous historical sites, buildings \\
        & Historical Artifacts & Museum collections, ancient relics \\
        & Historical Figures & Portraits of historical figures, statues \\
        & Architectural Styles & Traditional architecture, modern landmark buildings \\

        \midrule

        \multirow{5}{*}{\makecell{Natural Environment \\and Local Resources}} 
        & Food & Local specialty dishes, traditional festival foods \\  
        & Plants & Unique flowers, crops in a certain area \\  
        & Animals & Unique wild animals, livestock in a certain area \\  
        & Natural Scenery and Ecosystems & Unique natural landscapes, ecological reserves \\  
        & Markets and Shopping Traditions & Local markets, specialty shops \\
        
        \bottomrule  
    \end{tabular}  
    }
    \caption{Collected concepts, their overall categories, and descriptions in the dataset.}
    \label{tb-concept-category}
\end{table*}

\section{Experimental Details and Results}
\label{sec-app-exp}

\subsection{Experiment Setup}
\label{app:experiment_setup}

\noindent \textbf{Evaluation Models.}
We conduct evaluations on the following models: (a) open-source models including LLaVA-1.5~\citep{liu2024improved}, LLaVA-1.6-Mistral-7B-Instruct~\citep{liu2024visual}, LLaVA-OneVision~\citep{li2024llava}, LLaMA-3.2-Vision~\citep{Meta-llama-3.2}, Qwen2-VL~\citep{wang2024qwen2}, InternVL-2~\citep{chen2023internvl}, Phi-3-Vision~\citep{abdin2024phi}, MiniCPM-Llama3-V-2.5~\citep{yao2024minicpm}, GLM-4V~\citep{glm2024chatglm}, Pixtral 12B~\citep{agrawal2024pixtral}; (b) proprietary models such as GPT-4o~\citep{GPT-4o}, Gemini-1.5-Pro~\citep{team2023gemini}.

\noindent \textbf{Evaluation Setup.}
For LLaVA-1.5, LLaMA-3.2-Vision and InternVL-2, we use lmdeploy~\cite{2023lmdeploy} for inference acceleration. For other models, we use vllm~\cite{kwon2023efficient} for the acceleration of inference. For models before and after fine-tuning, we use the same acceleration toolkit to prevent potential impact. The number of questions differs across the three tasks. This is because generating questions for cultural knowledge and scenario reasoning is more complex, and in some cases, GPT-4o refused to provide answers, making it impossible to generate valid questions. For image recognition questions, we directly use the questions and options as prompts. For cultural knowledge and scene reasoning questions, we employ stepwise reasoning prompts to facilitate the reasoning explanation. The prompts are available in the Appendix~\ref{Prompt_List}. 
For all proprietary models, we utilize the default hyper-parameters, setting the temperature to 0 and the max tokens to 1,024. For all open-source models, $do\_sample$ is set to False, $max\_gen\_len$ is set to 512, and the temperature is set to 0.01.

\noindent \textbf{Training Setup.}
We use the official train ing scripts of LLaVA\footnote{\url{https://github.com/haotian-liu/LLaVA}.}, Phi\footnote{\url{https://github.com/microsoft/Phi-3CookBook}.}, and LLaMA\footnote{\url{https://github.com/meta-llama/llama-recipes/blob/main/recipes/quickstart/finetuning/finetune_vision_model.md}.} for model training, largely adhering to the original hyperparameters, except for appropriately adjusting the batch size to accommodate the GPU memory capacity. For LLaVA-1.5, a learning rate of \(2 \times 10^{-5}\) is used, with no weight decay applied (\(0.0\)). The learning rate followed a cosine schedule, gradually increasing during the initial phase with a warmup ratio of \(0.03\). For Phi-3, we use a learning rate of \(4 \times 10^{-5}\) and a weight decay of \(0.01\). A linear learning rate scheduler is utilized, with 50 warmup steps to stabilize the early training stage. For LLaMA-3.2, fine-tuning is conducted using a learning rate of \(1 \times 10^{-5}\) with no weight decay (\(0.0\)). A multiplicative learning rate decay is applied after each epoch, with a gamma value of \(0.85\). The batch sizes are set to 64, 16 and 32 respectively. All models are trained for one epoch on the training set and fully fine-tuned on 4×A100 80GB GPUs. For the training data, although we do not conduct large-scale human annotation, we synthesize the data using only concepts that passed either GPT-4o or human quality assurance, significantly improving the accuracy of the dataset. The prompts used for GPT quality check can be found in the Appendix \ref{Prompt_List}.

\subsection{Detailed Main Results}
\label{app-results-main}
Detailed results on different tasks, continents, and cultural categories can be found in \tablename~\ref{tab:zero-shot}.

\begin{table*}[htbp!]
    \centering
    \resizebox{0.96\textwidth}{!}{
    \begin{tabular}{l|ccc|ccccc|ccc}
        \toprule
        \multirow{3}{*}{\textbf{Model}} & \multicolumn{3}{c|}{\textbf{Task}} & \multicolumn{5}{c|}{\textbf{Continent}} & \multicolumn{3}{c}{\textbf{Category}} \\
        \cmidrule(){2-12} 
         & \makecell{Image \\ Recognition } & \makecell{Cultural \\ Knowledge} & \makecell{Scene \\ Reasoning} & Africa & Asia & Europe & America & Oceania & CHT & HL & NELR \\  
        \midrule
        \rowcolor[gray]{0.88}
        \multicolumn{12}{c}{\textit{Open-Source Models}} \\
        \midrule
        \llava 7B & 51.19 & 48.13 & 72.44 & 55.62 & 53.78 & 57.24 & 63.27 & 60.41 & 58.73 & 56.27 & 58.75 \\
        \llava 13B & 53.05 & 38.51 & 61.40 & 49.83 & 47.23 & 51.52 & 57.53 & 54.51 & 52.21 & 50.28 & 53.24 \\
        \llavanext 7B & 44.48 & 57.03 & 80.61 & 58.80 & 57.67 & 60.03 & 65.61 & 63.72 & 62.78 & 58.77 & 63.22 \\
        \llavaone 0.5B & 45.38 & 36.40 & 65.33 & 47.80 & 48.24 & 48.23 & 51.48 & 53.59 & 51.19 & 48.38 & 48.79 \\
        \llavaone 7B & 64.50 & 61.98 & 78.94 & 66.00 & 66.53 & 68.59 & 72.20 & 70.63 & 68.50 & 67.98 & 70.17 \\
        InternVL2 2B & 41.86 & 40.99 & 64.22 & 48.10 & 47.01 & 48.73 & 52.67 & 50.46 & 50.22 & 48.15 & 50.24 \\
        InternVL2 8B & 58.40 & 65.34 & 85.43 & 65.58 & 66.89 & 70.32 & 73.94 & 71.92 & 69.47 & 69.34 & 70.05 \\
        Phi-3-vision 4B & 54.85 & 56.43 & 79.09 & 60.60 & 60.99 & 63.21 & 67.78 & 68.69 & 64.22 & 62.40 & 65.73 \\
        MiniCPM-V-2\_6 8B & 65.07 & 71.69 & 88.00 & 71.52 & 72.59 & 76.10 & 77.60 & 74.86 & 74.77 & 74.62 & 74.93 \\
        GLM-4V 9B & 68.59 & 66.03 & 82.25 & 69.38 & 69.34 & 73.41 & 76.67 & 73.20 & 71.86 & 72.45 & 72.02 \\
        Pixtral 12B & 64.43 & 72.56 & 89.82 & 71.79 & 73.04 & 76.12 & 79.41 & 77.44 & 75.56 & 74.39 & 79.02 \\
        Qwen2-VL 7B & 75.51 & 71.90 & 77.96 & 73.59 & 72.73 & 75.71 & 78.93 & 77.90 & 75.05 & 74.57 & 77.78 \\
        Qwen2-VL 72B & 72.73 & 73.20 & 83.05 & 76.62 & 73.87 & 75.71 & 80.66 & 80.11 & 78.30 & 74.69 & 79.84 \\
        LLaMA-3.2-Vision 11B & 71.87 & 73.72 & 87.86 & 74.55 & 76.62 & 77.57 & 80.90 & 79.28 & 78.12 & 77.24 & 79.04 \\
        \midrule
        \rowcolor[gray]{0.88}
        \multicolumn{12}{c}{\textit{Proprietary Models}} \\
        \midrule
        Gemini-1.5-Pro & 81.88 & 87.03 & 93.60 & 84.91 & 86.14 & 88.06 & 89.62 & 87.08 & 86.50 & 87.43 & 88.36 \\
        GPT-4o & 84.67 & 90.86 & 96.16 & 88.69 & 90.22 & 90.91 & 90.92 & 89.41 & 91.53 & 89.50 & 92.58 \\
        \bottomrule
    \end{tabular}
    }
    \caption{Zero-shot accuracy on open-source and proprietary models across three tasks, five continents, and three categories. CHT denotes \emph{Cultural Heritage and Traditions}; HL denotes \emph{History and Landmarks}; and NELR denotes \emph{Natural Environment and Local Resources}.}
    \label{tab:zero-shot}
\end{table*}

\subsection{Detailed Fine-tuning Results}
\label{app-results-finetuning}

\begin{figure}[htbp]
    \centering
    \includegraphics[width=0.34\linewidth]{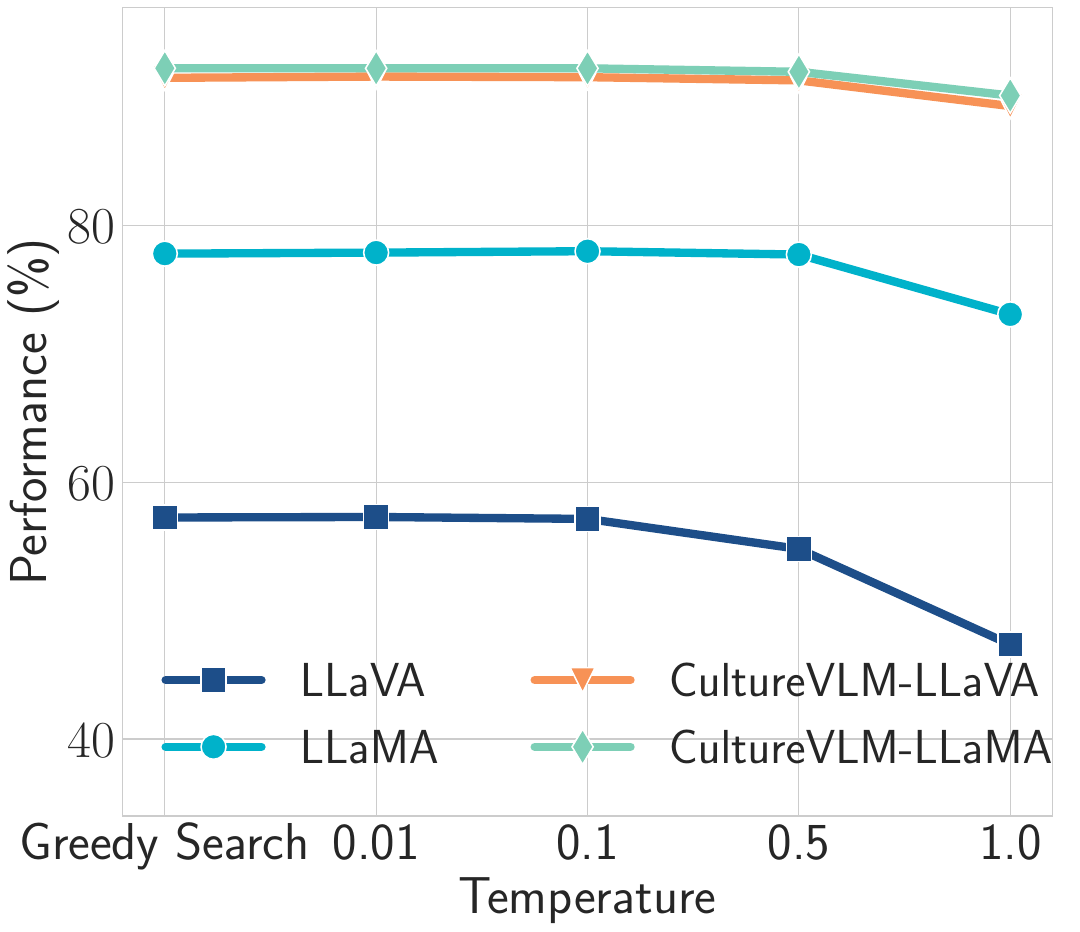}
    \caption{Impact of different decoding temperatures on performance}
    \label{fig:performance_temperature}
\end{figure}

The detailed results of the fine-tuned models are shown in Table \ref{tab:sft-performance}. Detailed results for different temperature settings can be found in \figurename~\ref{fig:performance_temperature}. Detailed results on the generalization of the fine-tuned model in different regions and for different categories can be found in \tablename~\ref{tb-acc-continents} and \tablename~\ref{tb-acc-categories}. Detailed results of the models before and after fine-tuning on the general VQA benchmark are shown in \tablename~\ref{tb-general-vqa}.

\begin{table*}[htbp!]
    \centering
    \resizebox{0.9\textwidth}{!}{
    \begin{tabular}{l|ccc|ccccc|ccc}
        \toprule
        \multirow{3}{*}{\textbf{Model}} & \multicolumn{3}{c|}{\textbf{Task}} & \multicolumn{5}{c|}{\textbf{Continent}} & \multicolumn{3}{c}{\textbf{Category}} \\
        \cmidrule(){2-12} 
         & \makecell{Image \\ Recognition } & \makecell{Cultural \\ Knowledge} & \makecell{Scene \\ Reasoning} & Africa & Asia & Europe & America & Oceania & CHT & HL & NELR \\  
        \midrule
        \llava 7B & 88.03 & 90.87 & 95.66 & 91.68 & 91.61 & 91.02 & 91.91 & 90.61 & 92.58 & 90.98 & 91.70 \\
        \llava 13B & 89.72 & 92.20 & 96.09 & 92.68 & 92.37 & 92.59 & 93.05 & 92.73 & 93.17 & 92.40 & 92.60 \\
        Phi-3-vision 4B & 87.53 & 90.84 & 95.77 & 90.80 & 91.56 & 91.41 & 90.91 & 91.16 & 92.83 & 90.50 & 92.35 \\
        LLaMA-3.2-Vision 11B & 89.13 & 91.49 & 96.20 & 91.99 & 92.24 & 91.82 & 93.08 & 91.34 & 93.20 & 91.78 & 92.50 \\
        \bottomrule
    \end{tabular}
    }
    \caption{Performance of fine-tuned models across three tasks, five continents, and three categories. CHT denotes \emph{Cultural Heritage and Traditions}; HL denotes \emph{History and Landmarks}; and NELR denotes \emph{Natural Environment and Local Resources}.}
    \label{tab:sft-performance}
\end{table*}

\begin{table}[t!]
\centering
\resizebox{.8\textwidth}{!}{
\begin{tabular}{ll|cccccc}
\toprule
\textbf{Task} & \textbf{Model} & \textbf{Total} & \textbf{Africa} & \textbf{Asia} & \textbf{Europe} & \textbf{America} & \textbf{Oceania} \\
\midrule
\multirow{5}{*}{Image Recognition} & LLaVA-v1.5-7B-Africa & 72.73 & 85.79 & 72.69 & 70.54 & 71.37 & 69.74 \\
      & LLaVA-v1.5-7B-America & 75.84 & 75.35 & 73.24 & 72.52 & 86.00 & 76.05 \\
      & LLaVA-v1.5-7B-Asia & 80.20 & 75.78 & 88.62 & 75.04 & 75.84 & 75.26 \\
      & LLaVA-v1.5-7B-Europe & 79.34 & 77.61 & 75.64 & 86.89 & 75.62 & 76.05 \\
      & LLaVA-v1.5-7B-Oceania & 64.51 & 64.05 & 63.14 & 63.60 & 65.28 & 83.95 \\
\midrule
\multirow{5}{*}{Cultural Knowledge} & LLaVA-v1.5-7B-Africa & 78.54 & 86.77 & 75.82 & 78.28 & 80.38 & 79.42 \\
      & LLaVA-v1.5-7B-America & 81.85 & 79.29 & 78.33 & 80.45 & 90.78 & 85.22 \\
      & LLaVA-v1.5-7B-Asia & 84.42 & 81.25 & 88.76 & 81.27 & 82.90 & 83.77 \\
      & LLaVA-v1.5-7B-Europe & 84.12 & 80.52 & 80.20 & 90.43 & 83.15 & 81.74 \\
      & LLaVA-v1.5-7B-Oceania & 73.78 & 73.16 & 72.38 & 73.91 & 75.03 & 81.16 \\
\midrule
\multirow{5}{*}{Scene Reasoning} & LLaVA-v1.5-7B-Africa & 91.29 & 93.63 & 91.06 & 90.87 & 91.41 & 91.41 \\
      & LLaVA-v1.5-7B-America & 92.70 & 93.40 & 92.24 & 91.85 & 94.81 & 91.69 \\
      & LLaVA-v1.5-7B-Asia & 94.16 & 94.44 & 95.11 & 92.92 & 94.52 & 92.52 \\
      & LLaVA-v1.5-7B-Europe & 93.26 & 92.71 & 92.64 & 94.08 & 93.45 & 92.52 \\
      & LLaVA-v1.5-7B-Oceania & 87.54 & 86.57 & 88.08 & 86.55 & 88.11 & 90.58 \\
\bottomrule
\end{tabular}}
\caption{Accuracy across different continents for each fine-tuned model.}
\label{tb-acc-continents}
\end{table}

\begin{table}[t!!]
\centering
\resizebox{.8\textwidth}{!}{
\begin{tabular}{ll|cccc}
\toprule
\textbf{Task} & \textbf{Model} & \textbf{Total} & \textbf{\makecell{Cultural Heritage \\ and Traditions}} & 
\textbf{\makecell{History and\\ Landmarks}} & 
\textbf{\makecell{Natural Environment \\ and Local Resources}} \\
\midrule
\multirow{3}{*}{Image Recognition} & LLaVA-v1.5-7B-CHT & 77.91 & 88.58 & 74.36 & 77.48 \\
      & LLaVA-v1.5-7B-HL & 84.76 & 81.57 & 86.89 & 80.25 \\
      & LLaVA-v1.5-7B-NELR & 73.46 & 74.89 & 69.18 & 89.88 \\
\midrule
\multirow{3}{*}{Cultural Knowledge}  & LLaVA-v1.5-7B-CHT & 83.29 & 88.53 & 81.71 & 82.16 \\
      & LLaVA-v1.5-7B-HL & 87.99 & 84.91 & 90.58 & 82.03 \\
      & LLaVA-v1.5-7B-NELR & 81.17 & 80.71 & 79.91 & 86.84 \\
\midrule
\multirow{3}{*}{Scene Reasoning}  & LLaVA-v1.5-7B-CHT & 92.96 & 95.27 & 92.45 & 91.73 \\
      & LLaVA-v1.5-7B-HL & 94.77 & 95.23 & 94.90 & 93.62 \\
      & LLaVA-v1.5-7B-NELR & 91.85 & 92.58 & 91.16 & 93.62 \\
\bottomrule
\end{tabular}}
\caption{Accuracy across different categories for each fine-tuned model.}
\label{tb-acc-categories}
\end{table}

\begin{table}[t!]
    \centering
    \resizebox{.4\textwidth}{!}{
    \begin{tabular}{ll|cc}
        \toprule
        \textbf{Dataset} & \textbf{Model} & \multicolumn{1}{c}{\textbf{Base}} & \multicolumn{1}{c}{\textbf{Finetuned}} \\ 
        \midrule
        \multirow{3}{*}{ScienceQA} & LLaVA 7B & 70.12 & 69.28 \\
        & LLaVA 13B & 74.91 & 75.04 \\
        & \llama 3.2 11B & 88.97 & 89.30 \\
        \midrule
        \multirow{3}{*}{TextVQA} & LLaVA 7B FT & 58.32 & 57.89 \\
        & LLaVA 13B FT & 61.18 & 61.13  \\
        & \llama 3.2 11B & 71.34  & 70.67 \\
        \bottomrule
    \end{tabular}
    }
    \caption{Performance of models before and after fine-tuning on general VQA datasets. \emph{Base} refers to the original model, while \emph{Finetuned} refers to the model adapted using our \dataset. The comparable performance across both versions suggests that finetuning on our dataset preserves the models’ natural language understanding and commonsense reasoning abilities.}
    \label{tb-general-vqa}
\end{table}

\section{Details of Human Annotation}
\label{sec-app-human}


\subsection{Statistics of Human Annotators and the Process}
\label{sec-app-human-stat}

\tablename~\ref{tb-stats-human} shows the statistics of human annotators in our study.
In total, through the contractor company, we hired $10$ expert annotators whose ages are between $20$ and $36$ with at least a bachelor's degree.
Most of them are within the non-AI areas such as education, specific languages, and history.
When assigning the annotation job, we asked each annotator to label the correctness, consistency, and relatedness of our questions and answers.
Specifically, correctness refers to the correctness of the generated questions and answers, consistency refers to the consistency between the questions, answers, and the concepts, and relatedness aims to make sure that the concepts and generated questions are related to each other.
We asked the annotators not only to make judgement based on their experience but also to manually check the results via Google search and other search engines.
Each instance is labeled by two experts and then verified by another to ensure correctness.
All annotation operations are performed following local laws and regulations to ensure fairness, equity, and accountability.

\begin{table}[t!]
\centering
\resizebox{.7\textwidth}{!}{
\begin{tabular}{cc|cc|c}
\toprule
\multicolumn{1}{c}{\textbf{Age}} & \% & \multicolumn{1}{c}{\textbf{Degree}} & \% & \multicolumn{1}{c}{\textbf{Major}}                                                                                                                    \\ \midrule
\multicolumn{1}{l}{20-25} & 50\% & \multicolumn{1}{l}{Bachelor} & 50\% & \multirow{2}{*}{\begin{tabular}[c]{@{}c@{}}Education; Specific languages; Computer science;   \\ Communication; Public relation; History\end{tabular}} \\ 
\multicolumn{1}{l}{26-36} & 50\% & \multicolumn{1}{l}{Master}   & 50\% &                                                                              \\ \bottomrule
\end{tabular}
}
\caption{Statistics of the human annotators to validate \dataset.}
\label{tb-stats-human}
\end{table}

\subsection{Accuracy of Human Annotation}
\label{sec-app-acc-human}

\tablename~\ref{tb-annotation-accuracy} shows the precision of human annotators in our generated questions. We then filter out all the wrong questions and only retain the correct ones. It is surprising that automatically generated questions can achieve high accuracy, indicating the promising future of the adoption of advanced AI models like GPT-4o for data collection and annotation.

\begin{table}[ht]
\centering
\resizebox{.5\textwidth}{!}{
\begin{tabular}{lc}
\toprule
\textbf{Check Item} & \textbf{Accuracy} \\ \midrule
Concept-Region Alignment & 98.25\% \\ 
Concept-Image Alignment & 99.49\% \\ 
Image Recognition Question-Answer Correctness & 98.61\% \\ 
Cultural Knowledge Question-Answer Correctness & 96.52\% \\ 
Scene Reasoning Question-Answer Correctness & 93.18\% \\ \bottomrule
\end{tabular}
}
\caption{Accuracy of \dataset based on the human annotations.}
\label{tb-annotation-accuracy}
\end{table}

\section{Case Study}
\label{sec-app-case-stydy}
Below, we present examples from \dataset representing three different countries. Each example includes a cultural concept, country, category, image recognition question, cultural knowledge question, and scene reasoning question.

\begin{mybody}
\begin{minipage}[t]{0.2\columnwidth}
\includegraphics[width=\linewidth, keepaspectratio]{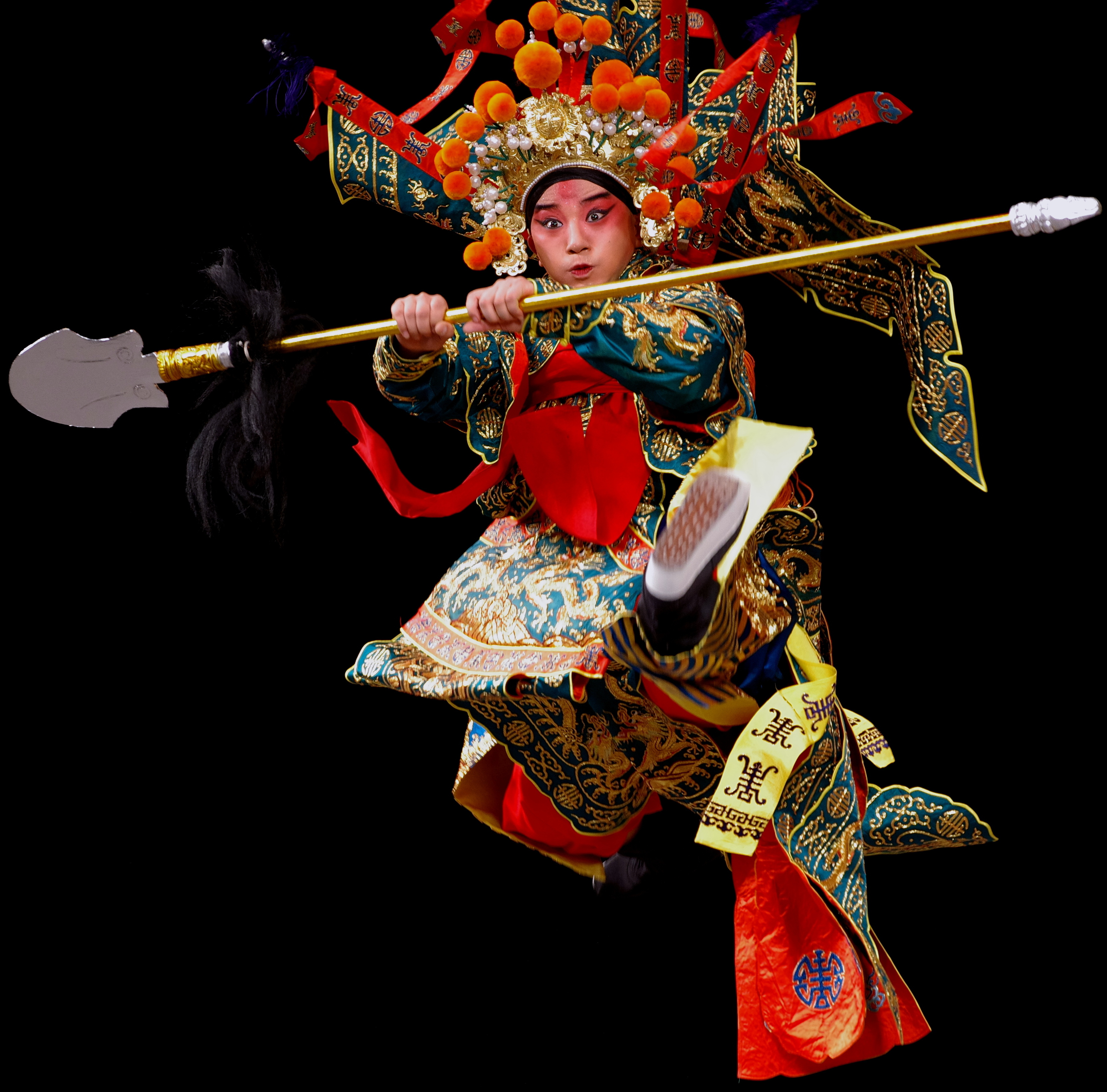}
\end{minipage}
\hspace{1mm}
\begin{minipage}[t!]{0.7\columnwidth}
\textbf{Concept:} Peking Opera \\
\textbf{Country:} China \\
\textbf{Category:} Music and Dance \\\\
\textbf{\textcolor{blue}{Image Recognition:}}\\
\textbf{Question:} What traditional Chinese performance is shown in this image? \\
\textbf{Options:} (A) Henan Opera (B) Peking Opera (C) Kunqu Opera (D) Yue Opera \\
\textbf{Ground truth:} \textbf{\green{(B) Peking Opera}} \\\\
\textbf{\textcolor{blue}{Cultural Knowledge:}}\\
\textbf{Question:} Which musical instrument is traditionally associated with accompanying the theatrical art form depicted in the image? \\
\textbf{Options:} (A) Guzheng (B) Pipa (C) Jinghu (D) Dizi \\
\textbf{Ground truth:} \textbf{\green{(C) Jinghu}} \\\\
\textbf{\textcolor{blue}{Scene Reasoning:}}\\
\textbf{Question:} In a theatrical performance known for its vibrant costumes and symbolic gestures, what might the intricate face paint and elaborate attire of a character symbolize? \\
\textbf{Options:} (A) The character's age and wisdom (B) The character's social status and role (C) The weather conditions in the story (D) The importance of technology in the narrative \\
\textbf{Ground truth:} \textbf{\green{(B) The character's social status and role}}
\end{minipage}
\end{mybody}

\begin{mybody}
\begin{minipage}[t]{0.2\columnwidth}
\includegraphics[width=\linewidth, keepaspectratio]{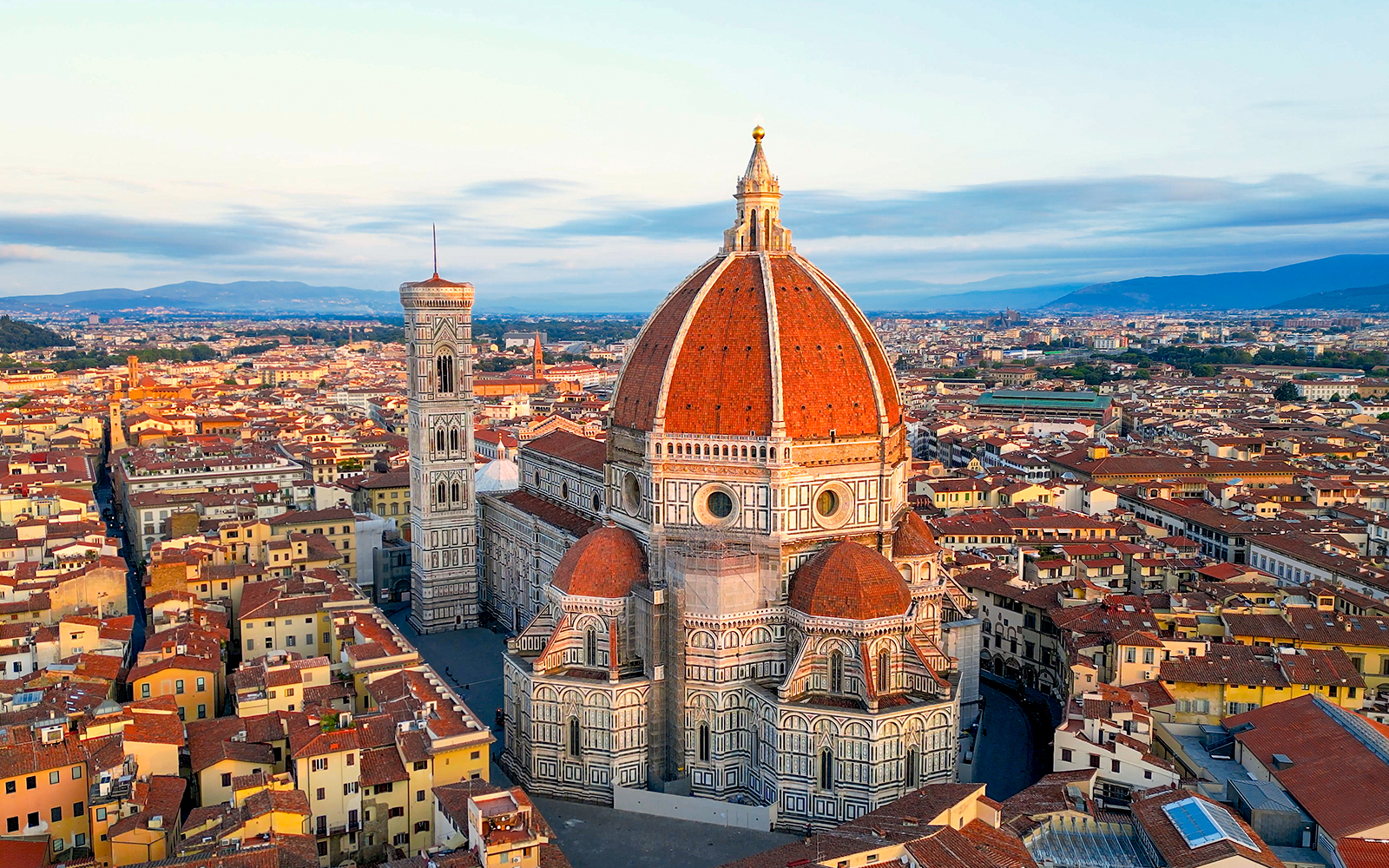}
\end{minipage}
\hspace{1mm}
\begin{minipage}[t!]{0.7\columnwidth}
\textbf{Concept:} Florence Cathedral \\
\textbf{Country:} Italy \\
\textbf{Category:} Famous Landmarks \\\\
\textbf{\textcolor{blue}{Image Recognition:}}\\
\textbf{Question:} What famous cathedral is shown in this image? \\
\textbf{Options:} (A) St. Peter's Basilica (B) Milan Cathedral (C) Florence Cathedral (D) Siena Cathedral \\
\textbf{Ground truth:} \textbf{\green{(C) Florence Cathedral}} \\\\
\textbf{\textcolor{blue}{Cultural Knowledge:}}\\
\textbf{Question:} Which renowned architect was responsible for engineering the innovative dome of the structure shown in the image? \\
\textbf{Options:} (A) Leon Battista Alberti (B) Filippo Brunelleschi (C) Giorgio Vasari (D) Michelangelo Buonarroti \\
\textbf{Ground truth:} \textbf{\green{(B) Filippo Brunelleschi}} \\\\
\textbf{\textcolor{blue}{Scene Reasoning:}}\\
\textbf{Question:} Imagine you are visiting the city depicted in the image during the Renaissance period. You hear local legends about a stone head on a famous structure there. According to the legend, what was the purpose of this stone head? \\
\textbf{Options:} (A) To ward off evil spirits (B) To grant wishes to passersby (C) To commemorate a donor who contributed a bell (D) To mark the entrance for pilgrims \\
\textbf{Ground truth:} \textbf{\green{(C) To commemorate a donor who contributed a bell}}
\end{minipage}
\end{mybody}

\begin{mybody}
\begin{minipage}[t]{0.2\columnwidth}
\includegraphics[width=\linewidth, keepaspectratio]{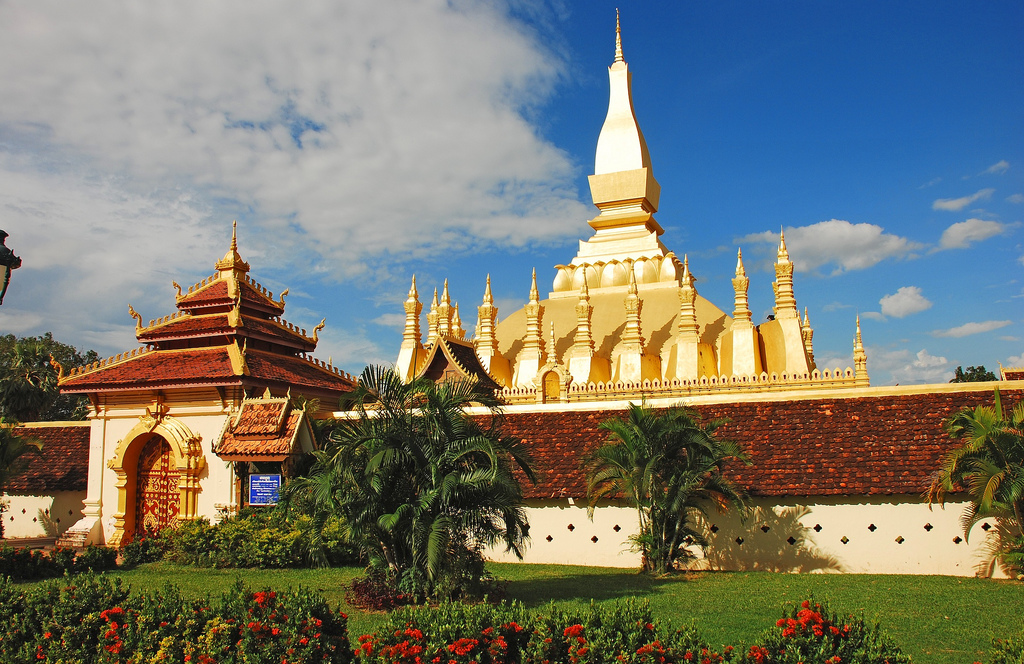}
\end{minipage}
\hspace{1mm}
\begin{minipage}[t!]{0.7\columnwidth}
\textbf{Concept:} Pha That Luang \\
\textbf{Country:} Lao People's Democratic Republic \\
\textbf{Category:} Famous Landmarks \\\\
\textbf{\textcolor{blue}{Image Recognition:}}\\
\textbf{Question:} What famous structure is shown in this image? \\
\textbf{Options:} (A) Wat Arun (B) Angkor Wat (C) Pha That Luang (D) Shwedagon Pagoda \\
\textbf{Ground truth:} \textbf{\green{(C) Pha That Luang}} \\\\
\textbf{\textcolor{blue}{Cultural Knowledge:}}\\
\textbf{Question:} The structure shown in the image, originally built as a Hindu temple, underwent major reconstruction during the reign of which historical figure when Buddhism gained prominence in the region? \\
\textbf{Options:} (A) King Jayavarman VII (B) King Setthathirath (C) King Anawrahta (D) King Ramkhamhaeng \\
\textbf{Ground truth:} \textbf{\green{(B) King Setthathirath}} \\\\
\textbf{\textcolor{blue}{Scene Reasoning:}}\\
\textbf{Question:} Imagine you are attending a festival at the site shown in the image, which is considered one of the most significant Buddhist celebrations in the country. During this event, which of the following cultural practices would you most likely observe that emphasizes the religious and national significance of this site? \\
\textbf{Options:} (A) A parade featuring traditional music and dance (B) A ceremony honoring ancient Hindu deities (C) A cooking contest of traditional Lao dishes (D) A fireworks display celebrating the lunar new year \\
\textbf{Ground truth:} \textbf{\green{(A) A parade featuring traditional music and dance}}
\end{minipage}
\end{mybody}

\section{Prompt List}
\label{Prompt_List}

\lstset{
    backgroundcolor=\color[RGB]{245,245,245},
    breaklines=true,
    breakindent=0pt,
    basicstyle=\ttfamily\small,
    frame=trbl,
    frameround = tttt,
    captionpos=b  
}




\begin{lstlisting}[caption={Prompt of concept judgement.}]
Concept: {concept}
Country: {country}
Class: {class_}

Please determine whether "{concept}" is a kind of {class_} that can reflect {country} culture and whether it can serve as a symbol of {country} culture (unique and very famous within {country}, and not commonly seen in other parts of the world).
Additionally, the symbol should not be a broad category that includes various specific items, but rather a distinct and indivisible entity, such as a specific dance form, a famous individual's photograph, a renowned landmark, etc.

Here are some counterexamples:
- Broad concepts like "fast food" (which includes burgers, fries, etc.), "traditional Chinese instruments" (which include guzheng, erhu, etc.) "Dance" (which include jazz dance, Square dancing, etc.) are not acceptable due to their lack of a unified visual marker.
- Concept itself is the wrong word, such as "...", "N/A".
- Concept exists in many regions, such as "pork", "duck" or "grapes" (which might be a specialty or staple in certain countries but is quite common in many places. However, specific concepts like "peking duck" or "schweinshaxe" would be correct). 

Please provide your short explanation and include your answer (Yes/No) into <<<>>>. For example, if you think "{concept}" is a specific and indivisible symbol of {country} culture, please write <<<Yes>>>.
\end{lstlisting}

\begin{lstlisting}[caption={Prompt for extracting cultural concept entities from Wikipedia documents.}]
Please extract cultural elements from the given wikipedia document that can represent {country} culture or are very famous in {country}, including the following categories:

### Categories
- Food: e.g., local specialty dishes, traditional festival foods.
- Plants: e.g., unique flowers, crops in {country}.
- Animals: e.g., unique wild animals, livestock in {country}.
- Famous Landmarks: e.g., famous historical sites, buildings.
- Festivals: e.g., unique festival celebration scenes.
- Historical Artifacts: e.g., museum collections, ancient relics.
- Historical Figures: e.g., portraits of historical figures, statues.
- Traditional Clothing: e.g., ethnic clothing, festival attire.
- Architectural Styles: e.g., traditional architecture, modern landmark buildings.
- Handicrafts and Artifacts: e.g., ethnic handicrafts, traditional handmade items.
- Music and Dance: e.g., traditional musical instruments, dance scenes.
- Religion and Belief: e.g., temples and churches, religious ceremonies.
- Natural Scenery and Ecosystems: e.g., unique natural landscapes, ecological reserves.
- Markets and Shopping Traditions: e.g., local markets, specialty shops.
- Entertainment and Performing Arts: e.g., theater performances, street performers.

Please note that not all categories may be included in the document. Only list the most famous cultural elements, with a total not exceeding 10. For categories without famous elements, use 'NA' to indicate. Directly output in the format "Category: [Element1, Element2, ...]", for example:

### Cultural Elements
- Food: [Food 1], [Food 2], ...
- Plants: NA
- Music and Dance: [Musical Instrument 1], [Dance Scene 2], ...

The following is the wikipedia document.
### Wikipedia Document
{wikipedia}
\end{lstlisting}

\begin{lstlisting}[caption={Prompt for generating scene recognition questions.}]
This is the {concept} of {country}. Your task is to generate a multiple-choice question that asks the user to identify what is shown in the image. Use the following format for your question:
- Question: [Your Question] Options: (A) [Option 1] (B) [Option 2] (C) [Option 3] (D) [Option 4]

For example, if the image shows a Peking Duck of China, the question and options should look like this:
- Question: What traditional dish is shown in this image? Options: (A) Cantonese Roast Duck (B) Peking Duck (C) Sichuan Spicy Duck (D) Nanjing Salted Duck

If the image shows the Erhu of China:
- Question: What musical instrument is shown in this image? Options: (A) Pipa (B) Erhu (C) Sanxian (D) Yangqin

If the image shows the White House of US:
- Question: What famous American building is shown in this image? Options: (A) The Capitol (B) The White House (C) The Lincoln Memorial (D) The Supreme Court Building

Ensure that "{concept}" should be included in one of the options. Ensure that the options are plausible but only one is the correct answer. The incorrect options should be similar enough to the correct one to create a challenge, but not so similar that they cause any potential ambiguity.
\end{lstlisting}

\begin{lstlisting}[caption={Prompt for generating the introduction of cultural concept.}]
Please provide a detailed introduction of {concept} of {country}, including information such as: Location and Features (where it is found or originates from, what makes it unique); Time period (when it was created or became significant); History (historical background and development, or any significant events); Cultural significance (cultural Context in {country}, modern-day significance); Stories or Legends (Any stories, legends, or folklore associated with {concept}).

Use the following format for your introduction:
Introduction of {concept}: [Detailed Introduction of {concept}]

Here are two examples:

Introduction of Peking Duck: Peking Duck is a famous Chinese dish that originated in Beijing during the Imperial era. The dish dates back to the Yuan Dynasty (1271-1368) and became a staple in the Ming Dynasty (1368-1644). Traditionally, Peking Duck is known for its thin, crispy skin and is served with pancakes, hoisin sauce, and scallions. The preparation involves inflating the duck to separate the skin from the fat, marinating it, and roasting it in a closed or hung oven. It is considered a national dish of China and a symbol of Chinese culinary art.

Introduction of The White House: The White House, located at 1600 Pennsylvania Avenue NW in Washington, D.C., is the official residence and workplace of the President of the United States. Construction began in 1792 and was completed in 1800. The building was designed by Irish-born architect James Hoban in the neoclassical style. It has been the residence of every U.S. president since John Adams. The White House has undergone several renovations and expansions, including the addition of the West Wing and the Oval Office. It is a symbol of the U.S. government and a site of significant historical events.

Now please provide the introduction for {concept}.
Introduction of {concept}:
\end{lstlisting}

\begin{lstlisting}[caption={Prompt for generating cultural knowledge questions based on the introduction of cultural concepts.}]
This public image shows the "{concept}" of {country}. Generate a multiple-choice question based on this image and the introduction of {concept}. Provide the correct answer immediately following the question.

Ensure the question delves into deeper cultural knowledge but does not directly name the {concept}. The options should be somewhat confusing to increase the difficulty, but there must be only one correct answer. Users can only answer based on the image, so don't mention any "introduction" or "{concept}" in the question. Use the following format for your generated question:
- Question: [Your Question] Options: (A) [Option 1] (B) [Option 2] (C) [Option 3] (D) [Option 4]
- Answer: (X) [Option X]

Here are two examples:

Image: Peking Duck
Introduction of Peking Duck: Peking Duck is a famous Chinese dish that originated in Beijing during the Imperial era. The dish dates back to the Yuan Dynasty (1271-1368) and became a staple in the Ming Dynasty (1368-1644). Traditionally, Peking Duck is known for its thin, crispy skin and is served with pancakes, hoisin sauce, and scallions. The preparation involves inflating the duck to separate the skin from the fat, marinating it, and roasting it in a closed or hung oven. It is considered a national dish of China and a symbol of Chinese culinary art.

- Question: During which dynasty did the dish shown in the image become a staple in the cuisine of its country? Options: (A) Tang Dynasty (B) Song Dynasty (C) Ming Dynasty (D) Qing Dynasty
- Answer: (C) Ming Dynasty


Image: The White House
Introduction of The White House: The White House, located at 1600 Pennsylvania Avenue NW in Washington, D.C., is the official residence and workplace of the President of the United States. Construction began in 1792 and was completed in 1800. The building was designed by Irish-born architect James Hoban in the neoclassical style. It has been the residence of every U.S. president since John Adams. The White House has undergone several renovations and expansions, including the addition of the West Wing and the Oval Office. It is a symbol of the U.S. government and a site of significant historical events.

- Question: Who was the architect responsible for designing the building shown in the image? Options: (A) James Hoban (B) Benjamin Latrobe (C) Thomas Jefferson (D) Charles Bulfinch
- Answer: (A) James Hoban


Now please generate the question for the Image: {concept} of {country}
{introduction}
\end{lstlisting}

\begin{lstlisting}[caption={Prompt for generating scene reasoning questions based on the introduction of cultural concepts.}]
This public image shows the "{concept}" of {country}. Generate a visual reasoning multiple-choice question based on this image and the introduction of {concept}. Provide the correct answer and reason immediately following the question.

Here are some requirements:
- The question must describe a specific scenario crafted to test deeper cultural understanding without directly naming {concept}. The scenario can be related to cultural background, regional characteristics, historical legends, or etiquette and customs, etc.
- The question needs to be related to the image but does not need to describe the content of the image.
- Ensure the question requires the user to recognize the image and use relevant knowledge to answer through reasoning based on the scenario provided. Users can only answer based on the image, so don't mention any "introduction" or "{concept}" in the question.

Use the following format for your introduction and question:
- Question: [Your Scenario-based Question] Options: (A) [Option 1] (B) [Option 2] (C) [Option 3] (D) [Option 4]
- Answer: (X) [Option X]
- Reason: [Your Reason for the Answer]

Here are two examples:

Image: Peking Duck
Introduction of Peking Duck: Peking Duck is a famous Chinese dish that originated in Beijing during the Imperial era. The dish dates back to the Yuan Dynasty (1271-1368) and became a staple in the Ming Dynasty (1368-1644). Traditionally, Peking Duck is known for its thin, crispy skin and is served with pancakes, hoisin sauce, and scallions. The preparation involves inflating the duck to separate the skin from the fat, marinating it, and roasting it in a closed or hung oven. This meticulous process ensures the skin becomes crispy while the meat remains tender. Peking Duck is often carved in front of diners and served in three stages: the skin, the meat, and a broth made from the bones. It is considered a national dish of {country} and a symbol of Chinese culinary art. Peking Duck has also been a part of many state banquets and diplomatic events, symbolizing Chinese hospitality and culinary excellence.

- Question: During a state banquet featuring the dish in the image, which aspect of its presentation is most likely emphasized to symbolize Chinese culinary excellence and hospitality? Options: (A) The use of exotic spices (B) The serving of the duck with rice (C) The incorporation of seafood (D) The carving of the duck in front of diners
- Answer: (D) The carving of the duck in front of diners
- Reason: The traditional carving of Peking Duck in front of diners highlights the skill involved in its preparation and serves as a symbol of Chinese culinary excellence and hospitality.


Image: The White House
Introduction of The White House: The White House, located at 1600 Pennsylvania Avenue NW in Washington, D.C., is the official residence and workplace of the President of the United States. Construction began in 1792 and was completed in 1800. The building was designed by Irish-born architect James Hoban in the neoclassical style, featuring a white-painted Aquia Creek sandstone exterior. It has been the residence of every U.S. president since John Adams. The White House has undergone several renovations and expansions, including the addition of the West Wing, East Wing, and the Oval Office. The building's iconic appearance and historical significance make it a symbol of the U.S. government and a site of significant historical events. The White House has been the location of many important decisions, meetings with foreign dignitaries, and addresses to the nation. It also serves as a museum of American history, housing numerous artifacts and pieces of art. The White House is not only a residence but also a working office, with various staff members ensuring the smooth operation of the executive branch of the U.S. government.

- Question: During a critical diplomatic event, the President is scheduled to meet with several foreign dignitaries to discuss global climate initiatives. As depicted in the image, which room inside the building is most likely to be used for this high-level diplomatic meeting? Options: (A) The Lincoln Bedroom (B) The Oval Office (C) The White House Kitchen (D) The East Room
- Answer: (B) The Oval Office
- Reason: The Oval Office is traditionally used for important meetings and discussions, making it the most likely choice for a high-level diplomatic meeting with foreign dignitaries.


Now please generate the question for the Image: {concept} of {country}
{introduction}
\end{lstlisting}

\begin{lstlisting}[caption={Prompt for generating knowledge-based reasoning response to image recognition questions in training set.}]
(Hint: This image shows the {concept} of {country}.)

Here is a question about this image:
{question}

First, describe the image in detail and analyze its features. Then, analyze the characteristics of the four options and compare each one with the features of the image. Finally, provide your final answer. Please include your answer into (). For example, if you choose A, please write (A). 
\end{lstlisting}

\begin{lstlisting}[caption={Prompt for generating knowledge-based reasoning response to cultural knowledge / scene reasoning questions in training set.}]
(Hint: This image shows the {concept} of {country}. {introduction})

Here is a question about this image:
{question}

First, provide a detailed description and identification of the image, analyzing its features. Then, conduct a comprehensive analysis of the question and four options based on the Hint and your knowledge. Finally, present the final answer. 

Please refrain from explicitly mentioning the "Hint" in your response, as these are for your discreet knowledge and not provided by the question. Please include your answer into (). For example, if you choose A, please write (A).

Response:
\end{lstlisting}

\begin{lstlisting}[caption={Prompt for stepwise reasoning in the evaluation of CultureVerse.}]
Here is a question about this image:
Question: {question} Options: {options}

First, describe and identify the image. Then, analyze the question and all four options in detail. Finally, provide the answer, indicating your final choice in parentheses. For example, if you choose A, please write (A).
\end{lstlisting}
\begin{table*}[]
    \centering
    \resizebox{0.95\textwidth}{!}{
    \begin{tabular}{lr | lr | lr}
        \toprule
        Country & Concept & Country & Concept & Country & Concept \\
        \midrule
        \textbf{SUM of Concepts} & \textbf{19,682} & Sri Lanka & 90 & Samoa & 30 \\
        India & 1,430 & Democratic People's Republic of Korea & 90 & Turkmenistan & 27 \\
        United States of America & 1,411 & Saudi Arabia & 87 & Qatar & 26 \\
        Italy & 661 & Lithuania & 86 & Guyana & 26 \\
        China & 545 & Malta & 86 & Kuwait & 25 \\
        Mexico & 524 & Uzbekistan & 84 & Paraguay & 24 \\
        Japan & 522 & Algeria & 83 & Sudan & 24 \\
        Philippines & 465 & Lebanon & 79 & Angola & 24 \\
        Indonesia & 400 & Nigeria & 78 & Fiji & 24 \\
        France & 374 & Colombia & 78 & Seychelles & 23 \\
        Russian Federation & 328 & Austria & 77 & Lesotho & 22 \\
        Greece & 300 & Cyprus & 75 & Barbados & 21 \\
        Germany & 273 & Mongolia & 74 & Dominica & 21 \\
        Egypt & 272 & Cuba & 73 & Mauritius & 20 \\
        Armenia & 259 & Bosnia and Herzegovina & 72 & Maldives & 19 \\
        Australia & 254 & Ecuador & 71 & Niger & 18 \\
        Spain & 249 & Slovakia & 65 & Zambia & 18 \\
        Georgia & 245 & Iceland & 65 & Antigua and Barbuda & 17 \\
        Brazil & 239 & Luxembourg & 65 & Saint Lucia & 16 \\
        Canada & 231 & Iraq & 62 & Eswatini & 16 \\
        Thailand & 228 & Ghana & 61 & Bahrain & 15 \\
        Myanmar & 226 & Albania & 60 & Papua New Guinea & 15 \\
        Ireland & 220 & Uruguay & 60 & Kazakhstan & 15 \\
        Pakistan & 218 & Montenegro & 60 & Tuvalu & 14 \\
        Nepal & 216 & Uganda & 58 & Liechtenstein & 14 \\
        New Zealand & 209 & Chile & 56 & Côte d'Ivoire & 14 \\
        Portugal & 199 & Senegal & 56 & Suriname & 13 \\
        Ukraine & 197 & United Republic of Tanzania & 56 & Bahamas & 12 \\
        Malaysia & 189 & United Arab Emirates & 56 & Eritrea & 12 \\
        Bangladesh & 188 & Guatemala & 54 & Mozambique & 12 \\
        Peru & 188 & Latvia & 53 & Burundi & 11 \\
        Poland & 186 & Yemen & 52 & Marshall Islands & 11 \\
        Bulgaria & 182 & Afghanistan & 51 & Honduras & 11 \\
        \makecell[l]{United Kingdom of Great \\ Britain and Northern Ireland} & 176 & Belarus & 50 & San Marino & 11 \\
        Croatia & 176 & Benin & 50 & Liberia & 11 \\
        Serbia & 172 & Oman & 50 & Tajikistan & 11 \\
        Romania & 167 & Dominican Republic & 49 & Solomon Islands & 10 \\
        Ethiopia & 155 & Tunisia & 48 & Comoros & 9 \\
        Cambodia & 154 & Trinidad and Tobago & 48 & Nauru & 9 \\
        Netherlands & 151 & El Salvador & 47 & Vanuatu & 8 \\
        Denmark & 142 & Monaco & 47 & Kiribati & 8 \\
        Lao People's Democratic Republic & 141 & Mali & 46 & Timor-Leste & 8 \\
        South Africa & 138 & Kenya & 45 & Grenada & 8 \\
        Argentina & 134 & Estonia & 45 & Sierra Leone & 7 \\
        Republic of Korea & 132 & Haiti & 44 & Rwanda & 7 \\
        Czechia & 130 & Jamaica & 43 & Guinea & 7 \\
        Bhutan & 130 & Belize & 42 & Libya & 7 \\
        Azerbaijan & 129 & Plurinational State of Bolivia & 41 & Andorra & 7 \\
        Morocco & 126 & Congo & 40 & Gambia & 7 \\
        Norway & 120 & Namibia & 39 & Burkina Faso & 7 \\
        Sweden & 112 & Somalia & 39 & South Sudan & 5 \\
        Finland & 112 & Nicaragua & 38 & Gabon & 5 \\
        Israel & 109 & Kyrgyzstan & 36 & Togo & 4 \\
        Türkiye & 109 & Bolivarian Republic of Venezuela & 35 & Malawi & 4 \\
        Viet Nam & 109 & Madagascar & 35 & Saint Kitts and Nevis & 4 \\
        Hungary & 101 & Republic of Moldova & 34 & Djibouti & 4 \\
        Ethnic\_and\_religiou\_groups & 94 & Zimbabwe & 33 & Saint Vincent and the Grenadines & 3 \\
        North Macedonia & 94 & Jordan & 33 & Central African Republic & 3 \\
        Slovenia & 93 & Tonga & 32 & Chad & 2 \\
        Islamic Republic of Iran & 93 & Botswana & 32 & Mauritania & 2 \\
        Switzerland & 93 & Cameroon & 31 & Panama & 1 \\
        Singapore & 93 & Costa Rica & 31 & Federated States of Micronesia & 1 \\
        Belgium & 91 & Democratic Republic of the Congo & 31 & Equatorial Guinea & 1 \\
        \bottomrule
    \end{tabular}
    }
    \caption{Number of cultural concepts of different countries or regions}
    \label{tb-country-concept}
\end{table*}




\end{document}